\documentclass{article}

\usepackage{times}
\usepackage{graphicx} %
\usepackage{subfigure} 

\usepackage{natbib}

\usepackage{algorithm}
\usepackage{algorithmic}

\usepackage[accepted]{icml2016}

\icmltitlerunning{Beyond CCA: Moment Matching for Multi-View Models}

\newcommand{\ppp}{\textbf}
\usepackage{natbib}
\usepackage{multibib}
\newcites{sup}{Supplementary References} 

\usepackage{tikz}
\usepackage{tkz-base}
\usetikzlibrary{arrows}
\usetikzlibrary{fit}
\usetikzlibrary{decorations}

\usepackage{times}

\usepackage{graphicx} 
\usepackage{subfigure} 

\usepackage{algorithm}
\usepackage{algorithmic}
\usepackage{hyperref}
\usepackage{url}

\usepackage{color}
\definecolor{darkcerulean}{rgb}{0.03, 0.27, 0.49} 
\hypersetup{
    colorlinks=true, 
    linkcolor=darkcerulean,  
    citecolor=darkcerulean,  
    filecolor=darkcerulean,  
    urlcolor=darkcerulean,   
}
\definecolor{backcolor}{RGB}{188, 211, 229}
\newcommand{\piccolor}{backcolor!50}

\usepackage{amssymb,amsmath,amsthm}
\newtheorem{theorem}{Theorem}
\usepackage{bbm}

\newcommand{\one}{1}
\newcommand{\zero}{0}
\newcommand{\norm}[1]{\left\|#1\right\|}

\newcommand{\inner}[1]{\left\langle#1\right\rangle}
\newcommand{\innerp}[1]{\langle#1\rangle}
\newcommand{\Rra}[1]{\left(#1\right)}
\newcommand{\rbra}[1]{\left(#1\right)}
\newcommand{\sbra}[1]{\left[#1\right]}
\newcommand{\cbra}[1]{\left\{#1\right\}}
\newcommand{\wt}[1]{\widetilde{#1}}
\newcommand{\wh}[1]{\widehat{#1}}
\newcommand{\diag}{\mathrm{diag}}

\newcommand{\off}{\mathrm{Off}}
\def\argmin{\mathop{\rm arg\,min}\limits}
\newcommand{\R}{\mathbb{R}}
\newcommand{\ebb}{\mathbb{E}}       
\newcommand{\webb}{\wh{\ebb}}       
\newcommand{\var}{\mathrm{var}}     
       
\newcommand{\cov}{\mathrm{cov}}     
       
\newcommand{\cum}{\mathrm{cum}}     
       
\newcommand{\sumk}{\sum\mathop{}_{k}}

\newcommand{\sumn}{\sum\mathop{}_{n}}

\newcommand{\poi}{\mathrm{Poisson}}
\newcommand{\gam}{\mathrm{Gamma}}

\newcommand{\acal}{\mathcal{A}}
\newcommand{\bcal}{\mathcal{B}}
\newcommand{\ccal}{\mathcal{C}}
\newcommand{\ecal}{\mathcal{E}}
\newcommand{\ncal}{\mathcal{N}}
\newcommand{\tcal}{\mathcal{T}}
\newcommand{\ci}{\perp\!\!\!\perp}

\makeatletter
\newenvironment{mi}{%
  \begin{list}{-}{}
  
}{%
  \end{list}
}
\makeatother

\begin{document} 

\twocolumn[
\icmltitle{Beyond CCA: Moment Matching for Multi-View Models}

\icmlauthor{Anastasia Podosinnikova}{anastasia.podosinnikova@inria.fr}
\icmlauthor{Francis Bach}{francis.bach@inria.fr}
\icmlauthor{Simon Lacoste-Julien}{firstname.lastname@inria.fr}
\icmladdress{INRIA - \'Ecole normale sup\'erieure, Paris}

\icmlkeywords{cca, canonical correlation analysis, multi-view models, moment matching, joint diagonalization, cumulant tensors, cp tensor decomposition}

\vskip 0.3in
]

\begin{abstract}
\noindent
We introduce three novel semi-parametric extensions of probabilistic canonical correlation analysis with identifiability guarantees. We consider moment matching techniques for estimation in these models.
For that, by drawing explicit links between the new models and a discrete version of independent component analysis~(DICA), we first extend the~DICA cumulant tensors to the new discrete version of CCA. 
By further using a close connection with independent component analysis, we introduce generalized covariance matrices, which can replace the cumulant tensors in the moment matching framework, and, therefore, improve sample complexity and simplify derivations and algorithms significantly. 
As the tensor power method or orthogonal joint diagonalization are not applicable in the new setting, we use non-orthogonal joint diagonalization techniques for matching the cumulants. We demonstrate performance of the proposed models and estimation techniques on experiments with both synthetic and real datasets.  
\end{abstract}

\section{Introduction}
Canonical correlation analysis (CCA), originally introduced by \citet{Hot1936}, is a common statistical tool for the analysis of  multi-view data. Examples of such data include, for instance, representation of some text in two languages \citep[e.g.,][]{VinEtAl2003} or images aligned with text data \citep[e.g.,][]{HarEtAl2004,GonEtAl2014}.
Given two multidimensional variables (or datasets), CCA finds two linear transformations (factor loading matrices) that mutually maximize the correlations between the transformed variables (or datasets). 
Together with its kernelized version \citep[see, e.g.,][]{ChrSha2004,BacJor2002}, CCA has  a wide range of applications (see, e.g., \citet{HarEtAl2004} for an overview).

\citet{BacJor2005} provide a probabilistic interpretation of CCA: they show that the maximum likelihood estimators of a particular Gaussian graphical model, which we refer to as Gaussian CCA, is equivalent to the classical CCA by \citet{Hot1936}. 
The key idea of Gaussian CCA is to allow some of the covariance in the two observed variables to be  explained by a linear transformation of common independent sources, while the rest of the covariance of each view is explained by their own (unstructured) noises. Importantly, the dimension of the common sources is often significantly smaller than the dimensions of the observations and, potentially, than the dimensions of the noise.
Examples of applications and extensions of Gaussian CCA are the works by \citet{SocFei2010}, for mapping visual and textual features to the same latent space, and \citet{HagEtAl2008}, for  machine translation applications. 

Gaussian CCA is subject to some well-known unidentifiability issues, in the same way as the closely related factor analysis model~\citep[FA;][]{Bar1987,Bas1994} and its special case, the probabilistic principal component analysis model \citep[PPCA;][]{TipBis1999,Row1998}.
Indeed, as FA and PPCA are identifiable only up to multiplication by any rotation matrix, Gaussian CCA is only identifiable up to multiplication by any invertible matrix. 
Although this unidentifiability does not affect the predictive performance of the model, it does affect the factor loading matrices and hence the interpretability of the latent factors. In FA and PPCA, one can enforce additional constraints to recover unique factor loading matrices \citep[see, e.g.,][]{Mur2012}.
A notable identifiable version of FA is independent component analysis \citep[ICA;][]{Jut1987,JutHer1991,ComJut2010}.
One of our goals is to introduce identifiable versions of  CCA.

The main contributions of this paper are as follows. We first introduce  for the first time, to the best of our knowledge, three new formulations of CCA: \emph{discrete, non-Gaussian, and mixed}  (see Section~\ref{sec:models}). We then provide \emph{ identifiability guarantees} for the new models (see Section~\ref{sec:identifiability}). Then, in order to use a moment matching framework for estimation, we first derive a \emph{new set of cumulant tensors} for the discrete version of CCA (Section~\ref{sec:cumulants}). We further replace these tensors with their approximations by \emph{generalized covariance matrices} for all three new models (Section~\ref{sec:generalized-covariances}). Finally, as opposed to standard approaches, we use a particular type of \emph{non-orthogonal joint diagonalization algorithms} for extracting the model parameters from the cumulant tensors or their approximations (Section~\ref{sec:jd}).

\ppp{Models.}
The new CCA models are adapted to applications where one or both of the data-views are either counts, like in the bag-of-words representation for text, or continuous data, for instance, any continuous representation of images. A key feature of CCA compared to joint PCA is the focus on modeling the common variations of the two views, as opposed to modeling all variations (including joint and marginal ones).

\ppp{Moment matching.} Regarding parameter estimation, we use the method of moments, also known as ``spectral methods.'' It recently regained popularity as an alternative to other estimation methods for graphical models, such as approximate variational inference or MCMC sampling. 
Estimation of a wide range of models is possible within the moment matching framework: 
ICA \citep[e.g.,][]{CarCom1996,ComJut2010}, 
mixtures of Gaussians \citep[e.g.,][]{AroKan2005,HsuKak2013}, 
latent Dirichlet allocation and topic models \citep{AroEtAl2012,AroEtAl2013,AnaEtAl2012,PodEtAl2015}, 
supervised topic models \citep{WanZhu2014}, 
Indian buffet process inference \citep{TunSmo2014},
stochastic languages \citep{BalEtAl2014},
mixture of hidden Markov models \citep{SubEtAl2014},  
neural networks   \citep[see, e.g.,][]{AnaSed2015,JanEtAl2016},
and other models \citep[see, e.g.,][and references therein]{AnaEtAl2014}.

Moment matching algorithms for estimation in graphical models mostly consist of two main steps: (a) construction of moments or cumulants with a particular diagonal structure and (b) joint diagonalization of the sample estimates of the moments or cumulants to estimate the parameters. 

\ppp{Cumulants and generalized covariance matrices.} By using the close connection between ICA and CCA, we first derive in Section~\ref{sec:cumulants} the cumulant tensors for the discrete version of CCA from the cumulant tensors of a discrete version of ICA (DICA) proposed by \citet{PodEtAl2015}. Extending the ideas from the ICA literature~\citep{Yer2000,TodHer2013}, we further generalize in Section~\ref{sec:generalized-covariances} cumulants as the derivatives of the cumulant generating function. This allows us to replace cumulant tensors with ``generalized covariance matrices'', while preserving the rest of the framework. As a consequence of working with the second-order information only, the derivations and algorithms get significantly simplified and the sample complexity potentially improves.

\ppp{Non-orthogonal joint diagonalization.} When estimating model parameters, both CCA cumulant tensors and generalized covariance matrices for CCA lead to non-symmetric approximate joint diagonalization problems. Therefore, the workhorses of the method of moments in similar context --- orthogonal diagonalization algorithms, such as the tensor power method \cite{AnaEtAl2014},  and orthogonal joint diagonalization \citep{BunEtAl1993,CarSou1996} --- are not applicable. As an alternative, we use a particular type of non-orthogonal Jacobi-like joint diagonalization algorithms (see Section~\ref{sec:jd}). Importantly, the joint diagonalization problem we deal with in this paper is conceptually different from the one considered, e.g., by \citet{KulEtAl2015} (and references therein) and, therefore, the respective algorithms are not applicable here.

\section{Multi-view models}
\subsection{Extensions of Gaussian CCA}
\label{sec:models}
\ppp{Gaussian CCA.}
Classical CCA \citep{Hot1936} aims to find projections $D_1\in\R^{M_1\times K}$ and $D_2\in\R^{M_2\times K}$, of two observation vectors $x_1\in\R^{M_1}$ and $x_2\in\R^{M_2}$, each representing a data-view, such that the projected data, $D_1^{\top}x_1$ and $D_2^{\top}x_2$, are maximally correlated. Similarly to classical PCA, the solution boils down to solving a generalized SVD problem. 
The following probabilistic interpretation of CCA is well known \citep{Bro1979,BacJor2005,KlaEtAl2013}. 
Given that $K$ sources are i.i.d.~standard normal random variables, $\alpha\sim\ncal(0,I_K)$, the \emph{Gaussian CCA} model is given by
\\[-0.6em]
\begin{equation}\label{gcca}
\begin{aligned}
x_1\, |\, \alpha,\;\mu_1,\;\Psi_1 &\sim \ncal(D_1 \alpha + \mu_1, \; \Psi_1), \\ 
x_2 \, |\, \alpha,\;\mu_2,\;\Psi_2 &\sim \ncal(D_2 \alpha + \mu_2, \; \Psi_2), 
\end{aligned}
\end{equation}
\\[-0.4em]
where the matrices $\Psi_1\in\R^{M_1\times M_1}$ and $\Psi_2\in\R^{M_2\times M_2}$ are positive semi-definite.
Then, the maximum likelihood solution of~\eqref{gcca} coincides (up to permutation, scaling, and multiplication by any invertible matrix) with the classical CCA solution. 
The model~\eqref{gcca} is equivalent to
\\[-0.6em]
\begin{equation}\label{pcca}
\begin{aligned}
x_1 &= D_1 \alpha + \varepsilon_1,\\ 
x_2 &= D_2 \alpha + \varepsilon_2,
\end{aligned}
\end{equation}
\\[-0.6em]
where the noise vectors are normal random variables, i.e. $\varepsilon_1\sim \ncal(\mu_1,\Psi_1)$ and $\varepsilon_2 \sim \ncal(\mu_2,\Psi_2)$,
and the following independence assumptions are made:
\\[-0.7em]
\begin{equation}\label{indep}
\begin{aligned}
&\alpha_1,\dots,\alpha_K \; \mbox{ are mutually independent}, \\
&\alpha \ci \varepsilon_1,\;\varepsilon_2 \quad \mbox{and} \quad \varepsilon_1 \ci \varepsilon_2.
\end{aligned}
\end{equation}
\\[-0.7em]
The following three models are our novel semi-parametric extensions of Gaussian CCA~\eqref{gcca}--\eqref{pcca}.

\ppp{Multi-view models.}
The first new model follows 
by dropping the Gaussianity assumption on~$\alpha$,~$\varepsilon_1$, and~$\varepsilon_2$. In particular, the \emph{non-Gaussian CCA} model is defined as
\\[-0.7em]
\begin{equation}\label{ncca}
\begin{aligned}
x_1 &= D_1 \alpha + \varepsilon_1,\\
x_2 &= D_2 \alpha + \varepsilon_2,
\end{aligned}
\end{equation}
\\[-0.7em]
where, as opposed to~\eqref{pcca}, no assumptions are made on the sources $\alpha$ and the noise $\varepsilon_1$ and $\varepsilon_2$ except for the independence assumption~\eqref{indep}.

Similarly to
\citet{PodEtAl2015}, we further ``discretize'' non-Gaussian CCA~\eqref{ncca} by applying the Poisson distribution to each view (independently on each variable):
\\[-0.7em]
\begin{equation}\label{dcca}
\begin{aligned}
x_1\,|\,\alpha,\;\varepsilon_1 & \sim \poi (D_1\alpha + \varepsilon_1),\\
x_2\,|\,\alpha,\;\varepsilon_2 & \sim \poi (D_2\alpha + \varepsilon_2).
\end{aligned}
\end{equation}
\\[-0.7em]
We obtain the (non-Gaussian) \emph{discrete CCA} (DCCA) model, which is adapted to count data (e.g., such as word counts in the bag-of-words model of text). In this case, the sources $\alpha$, the noise $\varepsilon_1$ and $\varepsilon_2$, and the matrices $D_1$ and $D_2$ have non-negative components.

Finally, by combining non-Gaussian and discrete CCA, we also introduce the \emph{mixed CCA} (MCCA) model:
\\[-0.7em]
\begin{equation}\label{mcca}
\begin{aligned}
x_1 &=D_1\alpha + \varepsilon_1,\\
x_2\,|\,\alpha,\;\varepsilon_2 &\sim \poi (D_2\alpha + \varepsilon_2),
\end{aligned}
\end{equation}
\\[-0.7em]
which is adapted to a combination of discrete and continuous data (e.g., such as images represented as continuous vectors aligned with text represented as counts).  Note that no assumptions are made on distributions of the sources $\alpha$ except for independence~\eqref{indep}.

\setlength{\textfloatsep}{9pt}
\begin{figure}
\centering
\begin{tikzpicture}
[
observed/.style={minimum size=20pt,circle,draw=black,fill=\piccolor,inner sep=1pt, thick},
unobserved/.style={minimum size=17pt,circle,draw,inner sep=1pt, thick},
hyper/.style={minimum size=3pt,circle,fill=black},
arrow/.style={
    decoration={markings,mark=at position 1 with {\arrow[scale=2,#1]{>}}},
    postaction={decorate}, thick},
]
\node (alpha) [unobserved] at (0,0) {$\alpha_k$};
\node (betao) [unobserved] at (-2,0.2) {$\varepsilon_1$};
\node (betat) [unobserved] at (2, 0.2) {$\varepsilon_2$};
\node (xo)    [observed]   at (-1.2,-1.7) {$x_{1m}$};
\node (xt)    [observed]   at (1.2,-1.7) {$x_{2m}$};
\filldraw [black] (-1.0,-2.9) circle (2pt);
\node (Do) [label=below:$D_1$,inner sep=2pt, outer sep=0pt] at (-1.0, -2.9) {};
\filldraw [black] (1.4, -2.9) circle (2pt);
\node (Dt) [label=below:$D_2$,inner sep=2pt, outer sep=0pt] at (1.4, -2.9) {};

\path
(betao) edge [arrow] (xo)
(alpha) edge [arrow] (xo)
(alpha) edge [arrow] (xt)
(betat) edge [arrow] (xt)
(Do)    edge [arrow] (xo)
(Dt)    edge [arrow] (xt)
;

\node [draw,fit=(alpha),inner sep=9pt, outer sep=0pt, rounded corners=.3em, thick] (plate-alpha) {};
\node [below right,inner sep=1.5pt,outer sep=0pt] at (plate-alpha.north west) {$K$};
\node [draw,fit=(xo),inner sep=10pt, outer sep=0pt, rounded corners=.3em, thick] (plate-xo) {};
\node [above right,inner sep=1.1pt, outer sep=0pt] at (plate-xo.south west) {$M_1$};
\node [draw,fit=(xt),inner sep=10pt, outer sep=0pt, rounded corners=.3em, thick] (plate-xt) {};
\node [above right, inner sep=1.1pt, outer sep=0pt] at (plate-xt.south west) {$M_2$};
\node [draw,fit=(xo) (xt) (alpha) (betao) (betat) (plate-xo) (plate-alpha),inner sep=6pt,rounded corners=.3em, thick] (plate-doc) {};
\node [above right,inner sep=1.5pt, outer sep=0pt] at (plate-doc.south west) {$N$};
\end{tikzpicture}
\vspace*{-.5cm}
\caption{Graphical models for non-Gaussian~\eqref{ncca}, discrete~\eqref{dcca}, and mixed~\eqref{mcca} CCA.}
\label{fig:cca}
\end{figure}
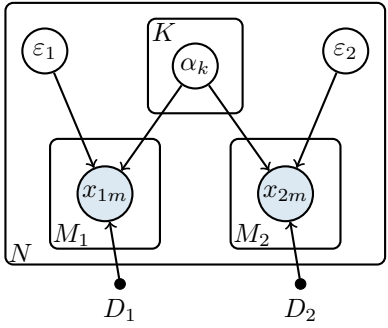

The plate diagram for the models~\eqref{ncca}--\eqref{mcca} is presented in Fig.~\ref{fig:cca}. 
We call $D_1$ and $D_2$  \emph{factor loading matrices} (see a comment on this naming convention in Appendix~\ref{app:naming_convention}).

\ppp{Relation between PCA and CCA.}
The key difference between Gaussian CCA and the closely related FA/PPCA models is that the noise in each view of Gaussian CCA is not assumed to be isotropic unlike for FA/PPCA. 
In other words, the components of the noise are not assumed to be independent or, equivalently, the noise covariance matrix does not have to be diagonal and may exhibit a strong structure.
In this paper, we never assume any diagonal structure of the covariance matrices of the noises of the models~\eqref{ncca}--\eqref{mcca}.
The following example illustrates the mentioned relation. Assuming  a linear structure for the noise, (non-) Gaussian CCA (NCCA) takes the form
\\[-0.6em]
\begin{equation}
\label{linear-cca}
\begin{aligned}
x_1 & = D_1\alpha + F_1 \beta_1, \\
x_2 & = D_2\alpha + F_2 \beta_2,
\end{aligned}
\end{equation}
\\[-0.4em]
where $\varepsilon_1 = F_1\beta_1$ with $\beta_1\in\R^{K_1}$ and $\varepsilon_2 = F_2\beta_2$ with $\beta_2\in\R^{K_2}$. By stacking the vectors on the top of each other
\\[-1.5em]
\begin{equation}
\label{stacking}
x = \begin{pmatrix} x_1 \\ x_2 \end{pmatrix}, \; D = \begin{pmatrix} D_1 & F_1 & \zero \\ D_2 & \zero & F_2 \end{pmatrix}, \; z = \begin{pmatrix} \alpha \\ \beta_1 \\ \beta_2 \end{pmatrix},
\end{equation}
\\[-1.0em]
we rewrite the model as
$x = D z$. Assuming that the noise sources $\beta_1$ and $\beta_2$ have mutually independent components, ICA is recovered. If the sources $z$ are further assumed to be Gaussian, $x=Dz$ corresponds to PPCA.
However, we do not assume the noise in Gaussian CCA (and in~\eqref{ncca}--\eqref{mcca}) to have a very specific low dimensional structure.

\ppp{Related work.}
Some extensions of Gaussian CCA were proposed in the literature: exponential family CCA \citep{Vir2010,KlaEtAl2010} and Bayesian CCA \citep[see, e.g.,][and references therein]{KlaEtAl2013}. 
Although exponential family CCA can also be discretized, it assumes in practice that the prior of the sources is a specific combination of Gaussians.
Bayesian CCA models the factor loading matrices and the covariance matrix of Gaussian CCA.
Sampling or approximate variational inference are used for estimation and inference in both models. Both models, however, lack our identifiability guarantees and are quite different from the models~\eqref{ncca}--\eqref{mcca}.
\citet{SonEtAl2014} consider a multi-view framework to deal with non-parametric mixture components, while our approach is semi-parametric with an explicit linear structure (our loading matrices) and makes the explicit link with CCA. See also~\citet{GeZou2016} for a related approach.

\subsection{Identifiability}
\label{sec:identifiability}
In this section, the identifiability of the factor loading matrices $D_1$ and $D_2$ is discussed. In general, for the type of models considered, the unidentifiability to permutation and scaling cannot be avoided. In practice, this unidentifiability is however easy to handle
and, in the following, we only consider identifiability up to permutation and scaling.

ICA can be seen as an identifiable analog of FA/PPCA.
Indeed, it is known that the mixing matrix $D$ of ICA is identifiable if at most one source is Gaussian \citep{Com1994}. The factor loading matrix of FA/PPCA is unidentifiable since it is defined only up to
multiplication by any orthogonal rotation matrix.

Similarly, the factor loading matrices of Gaussian CCA~\eqref{gcca}, which can be seen as a multi-view extension of PPCA, are identifiable only up to multiplication by any invertible matrix \citep{BacJor2005}. We show the identifiability results for the new models~\eqref{ncca}--\eqref{mcca}: the factor loading matrices of these models are identifiable if at most one source is Gaussian (see Appendix~\ref{app:identifiability} for a proof).

\vspace{0em}
\begin{theorem}
Assume that matrices $D_1\in\R^{M_1\times K}$ and $D_2\in\R^{M_2\times K}$, where $K \le \min(M_1,\;M_2)$, have full rank. If the covariance matrices $\cov(x_1)$ and $\cov(x_2)$ exist and if at most one source $\alpha_k$, for $k=1,\dots,K$, is Gaussian and none of the sources are deterministic, then the models~\eqref{ncca}--\eqref{mcca} are identifiable (up to scaling and joint permutation).
\label{thm:identif}
\vspace{-1.5em}
\end{theorem}

Importantly, the permutation unidentifiability does not destroy the alignment in the factor loading matrices, that is, for some permutation matrix $P$, if $D_1P$ is the factor loading matrix of the first view, than $D_2P$ must be the factor loading matrix of the second view. 
This property is important for the interpretability of the factor loading matrices and, in particular, is used in our experiments in 
 Section~\ref{sec:experiments}.

\section{The cumulants and generalized covariances}
In this section, we first derive the cumulant tensors for the discrete CCA model (Section~\ref{sec:cumulants}) and then generalized covariance matrices (Section~\ref{sec:generalized-covariances}) for the models~\eqref{ncca}--\eqref{mcca}. We show that both cumulants and generalized covariances have a special diagonal form and, therefore, can be efficiently used within the moment matching framework (Section~\ref{sec:jd}).

\subsection{From discrete ICA to discrete CCA}
\label{sec:cumulants}
In this section, we derive the DCCA  cumulants as an extension of the cumulants of discrete independent component analysis \citep[DICA;][]{PodEtAl2015}.

\ppp{Discrete ICA.}
\citet{PodEtAl2015}  consider the discrete ICA model~\eqref{dica}, where $x\in\R^M$ has conditionally independent Poisson components with mean $D\alpha$ and $\alpha\in\R^K$ has independent non-negative components:
\\[-0.8em]
\begin{equation}\label{dica}
\begin{aligned}
x\,|\,\alpha & \sim \poi(D\alpha).
\end{aligned}
\end{equation}
\\[-1.3em]
For estimating the factor loading matrix $D$, \citet{PodEtAl2015} propose an algorithm based on the moment matching method with the cumulants of the DICA model.
In particular, they define the DICA S-covariance matrix and T-cumulant tensor as
\\[-1.5em]
\begin{equation}\label{STdica}
\begin{aligned}
S &:= \cov(x) - \diag\sbra{ \ebb x}, \\
\sbra{T}_{m_1m_2m_3} &:= \cum(x)_{m_1m_2m_3} + [\tau]_{m_1m_2m_3},
\end{aligned}
\end{equation}
\\[-0.9em]
where 
indices $m_1$, $m_2$, and $m_3$ take the values in $1,\dots, M$, and 
$\sbra{ \tau }_{m_1m_2m_3}  = 2\delta_{m_1m_2m_3} \ebb x_{m_1} -\delta_{m_2m_3} \cov(x)_{m_1m_2} - \delta_{m_1m_3} \cov(x)_{m_1m_2} - \delta_{m_1m_2} \cov(x)_{m_1m_3}$
with
 $\delta$ being the Kronecker delta. 
For completeness, we outline the derivation by \citet{PodEtAl2015} below.
Let $y := D\alpha$. By the law of total expectation $\ebb(x) = \ebb(x|y) = \ebb(y)$ and by the law of total covariance
\\[-0.4em]
\vspace{-1.1em}
\begin{equation*}
\begin{aligned}
\cov(x) & = \ebb[ \cov(x|y) ] + \cov[ \ebb(x|y), \; \ebb(x|y) ] \\
& = \diag[ \ebb(y) ] + \cov(y),
\end{aligned}
\end{equation*}
\\[-0.8em]
since all the cumulants of a Poisson random variable with parameter $y$ are equal to $y$.
Therefore, $S = \cov(y)$. Similarly, by the law of total cumulance $T = \cum(y)$.
Then, by the multilinearity property for cumulants, one obtains
\\[-0.8em]
\begin{equation}\label{diagSTdica}
\begin{aligned}
S&=D\,\cov(\alpha)D^{\top}, \\
T&= \cum(\alpha) \times_1 D^{\top} \times_2 D^{\top} \times_3 D^{\top},
\end{aligned}
\end{equation}
\\[-0.8em]
where $\times_i$ denotes the $i$-mode tensor-matrix product
\citep[see, e.g.,][]{KolBad2009}.
Since the covariance $\cov(\alpha)$ and cumulant $\cum(\alpha)$ of the independent sources are diagonal,~\eqref{diagSTdica} is called
the \emph{diagonal form}.  
This diagonal form is further used for estimation of $D$ (see Section~\ref{sec:jd}).

\ppp{Noisy discrete ICA.}
The following noisy version~\eqref{ndica} of the DICA model reveals the connection between DICA and DCCA. 
Noisy discrete ICA is obtained by adding non-negative noise $\varepsilon$, such that $\alpha \ci \varepsilon$, to discrete ICA~\eqref{dica}:
\\[-0.8em]
\begin{equation}\label{ndica}
x\,|\,\alpha,\;\varepsilon \sim \poi\Rra{ D\alpha + \varepsilon}.
\end{equation}
\\[-1.2em]
Let $y:=D\alpha+\varepsilon$ and $S$ and $T$ are defined as in~\eqref{STdica}.
Then a simple extension of the derivations from above gives $S= \cov(y)$ and $T=\cum(y)$.
Since the covariance matrix (cumulant tensor) of the sum of two independent multivariate random variables, $D\alpha$ and $\varepsilon$, is equal to the sum of the covariance matrices (cumulant tensors) of these variables, the ``perturbed'' version of the diagonal form~\eqref{diagSTdica} follows
\\[-1.4em]
\begin{equation}\label{diagSTndica}
\begin{aligned}
S &= D\cov(\alpha) D^{\top} + \cov(\varepsilon), \\
T &= \cum(\alpha) \times_1 D^{\top} \times_2 D^{\top} \times_3 D^{\top} + \cum(\varepsilon).
\end{aligned}
\end{equation}
\\[-1.9em]

\ppp{DCCA cumulants.}
By analogy with~\eqref{stacking}, stacking the observations $x=[x_1; \; x_2]$, the factor loading matrices $D=[D_1; \; D_2]$, and the noise vectors $\varepsilon=[\varepsilon_1; \; \varepsilon_2]$ of discrete CCA~\eqref{dcca} gives a noisy version of discrete ICA with a particular form of the covariance matrix of the noise:
\\[-0.5em]
\begin{equation}\label{cov-noise-stacked-cca}
\textstyle
\cov(\varepsilon) = \begin{pmatrix} \cov(\varepsilon_1) & \zero \\ \zero & \cov(\varepsilon_2) \end{pmatrix},
\end{equation}
\\[-0.5em]
which is due to the independence $\varepsilon_1 \ci \varepsilon_2$.
Similarly, the cumulant $\cum(\varepsilon)$ of the noise has only two diagonal blocks which are non-zero. Therefore, considering only those parts of the S-covariance matrix and T-cumulant tensor of noisy DICA that correspond to zero blocks of the covariance $\cov(\varepsilon)$ and cumulant $\cum(\varepsilon)$, gives immediately a matrix and tensor with a diagonal structure similar to the one in~\eqref{diagSTdica}. Those blocks are the cross-covariance and cross-cumulants of $x_1$ and $x_2$.

We define the \emph{S-covariance matrix of discrete CCA}%
\footnote{ Note that $S_{21} := \cov(x_2,x_1)$ is just the transpose of $S_{12}$. 
}
as the cross-covariance matrix of $x_1$ and $x_2$:
\\[-1.1em]
\begin{equation}\label{Sdcca}
S_{12} := \cov(x_1,x_2).
\end{equation}
\\[-1.2em]
From~\eqref{diagSTndica} and~\eqref{cov-noise-stacked-cca}, the matrix $S_{12}$ has the following diagonal form
\\[-1.6em]
\begin{equation}\label{diagSdcca}
S_{12} = D_1 \cov(\alpha) D_2^{\top}.
\end{equation}
\\[-1.2em]
Similarly, we define the \emph{T-cumulant tensors of discrete CCA} (
$T_{121}\in\R^{M_1\times M_2\times M_1}$ and $T_{122}\in\R^{M_1\times M_2\times M_2}$) through the cross-cumulants of $x_1$ and $x_2$, for $j=1,2$:
\\[-1.3em]
\begin{equation}\label{Tdcca}
\begin{aligned}
\sbra{ T_{12j} }_{m_1m_2\tilde{m}_j} & := [\cum(x_1,x_2,x_j)]_{m_1m_2\tilde{m}_j} \\
& - \delta_{m_j\tilde{m}_j}[\cov(x_1,x_2)]_{m_1m_2},
\end{aligned}
\end{equation}
\\[-0.9em]
where the indices $m_1$, $m_2$, and $\tilde{m}_j$ take the values $m_1\in 1,\dots,M_1$, $m_2 \in 1,\dots,M_2$, and $\tilde{m}_j\in 1,\dots, M_j$.
From~\eqref{diagSTdica} and the mentioned block structure~\eqref{cov-noise-stacked-cca} of $\cov(\varepsilon)$, the DCCA T-cumulants have the diagonal form:
\\[-0.6em]
\begin{equation}\label{diagTdcca}
\begin{aligned}
T_{121} &= \cum(\alpha) \times_1 D_1^{\top} \times_2 D_2^{\top} \times_3 D_1^{\top}, \\
T_{122} &= \cum(\alpha) \times_1 D_1^{\top} \times_2 D_2^{\top} \times_3 D_2^{\top}.
\end{aligned}
\end{equation}
\\[-0.6em]
In Section~\ref{sec:jd}, we show how to estimate the factor loading matrices $D_1$ and $D_2$ using the diagonal form~\eqref{diagSdcca} and~\eqref{diagTdcca}.
Before that, in Section~\ref{sec:generalized-covariances}, we first derive the generalized covariance matrices of discrete ICA and the CCA models~\eqref{ncca}--\eqref{mcca}  as an extension of the ideas by \citet{Yer2000,TodHer2013}.

\subsection{Generalized covariance matrices}
\label{sec:generalized-covariances}
In this section, we introduce the generalization of the S-covariance matrix for both  DICA and the CCA models~\eqref{ncca}--\eqref{mcca},
which are obtained through the Hessian of the cumulant generating function. We show that (a) the generalized covariance matrices can be used for approximation of the T-cumulant tensors using generalized derivatives and (b) in the DICA case, these generalized covariance matrices have the diagonal form analogous to~\eqref{diagSTdica}, and, in the CCA case, they have the diagonal form analogous to~\eqref{diagSdcca}. 
Therefore, generalized covariance matrices can be seen as a substitute for the T-cumulant tensors in the moment matching framework. This (a) significantly simplifies derivations and the final expressions used for implementation of resulting algorithms and (b)  potentially improves the sample complexity, since only the second-order information is used.

\ppp{Generalized covariance matrices.} The idea of generalized covariance matrices is inspired by the similar extension of the ICA cumulants by \citet{Yer2000}. 

The cumulant generating function (CGF) of a multivariate random variable $x\in\R^M$ is defined as
\\[-0.9em]
\begin{equation}\label{cgf}
K_x(t) = \log { \ebb ( e^{t^{\top}x} ) },
\end{equation}
\\[-1.4em]
for $t\in\R^M$. The cumulants $\kappa_s(x)$, for $s=1,2,3,\dots$, are the coefficients of the Taylor series expansion of the CGF evaluated at zero. Therefore, the cumulants are the derivatives of the CGF evaluated at zero:
$\kappa_s(x) = \nabla^s K_x(0)$, $s=1,2,3,\dots$, where $\nabla^sK_x(t)$ is the $s$-th order derivative of $K_x(t)$ with respect to $t$. Thus, the expectation of $x$ is the gradient $\ebb(x) = \nabla K_x(0)$ and the covariance of $x$ is the Hessian $\cov(x)=\nabla^2 K_x(0)$ of the CGF evaluated at zero.

The extension of cumulants then follows immediately: for $t\in\R^M$, we refer to the derivatives $\nabla^sK_x(t)$ of the CGF as the \emph{generalized cumulants}. The respective parameter $t$ is called a \emph{processing point}. In particular, the gradient, $\nabla K_x(t)$, and Hessian, $\nabla^2 K_x(t)$, of the CGF are referred to as the \emph{generalized expectation} and \emph{generalized covariance matrix}, respectively:
\\[-1.3em]
\begin{align}
\textstyle
\label{gen-expectation}
\textstyle
\ecal_x(t) &:= \nabla K_x(t) = \frac{\ebb ({ x e^{t^{\top}x}})}{\ebb ( e^{t^{\top}x} ) }, \\[-0.1em]
\textstyle
\label{gen-covariance}
\ccal_x(t) &:= \nabla^2 K_x(t) = \frac{\ebb (xx^{\top} e^{t^{\top}x}) }{\ebb ( e^{t^{\top}x} ) } - \ecal_x(t) \ecal_x(t)^{\top}.
\end{align}
\\[-1.3em] 
We now outline the key ideas of this section. When a multivariate random variable $\alpha\in\R^K$ has independent components, its CGF $K_{\alpha}(h)=\log { \ebb(e^{h^{\top}\alpha}) }$, for some $h\in\R^K$, is equal to a sum of decoupled terms:
$
K_{\alpha}(h) = \sum_k \log \ebb(e^{h_k\alpha_k}).
$
Therefore, the Hessian $\nabla^2K_{\alpha}(h)$ of the CGF $K_{\alpha}(h)$ is diagonal (see Appendix~\ref{app:gen-cums-alpha}). Like   covariance matrices, these Hessians (a.k.a.~generalized covariance matrices) are subject to the multilinearity property for linear transformations of a vector, hence the resulting diagonal structure of the form~\eqref{diagSTdica}. This is essentially the previous ICA work \citep{Yer2000,TodHer2013}.
Below we generalize these ideas first to the discrete ICA case and then to the CCA models~\eqref{ncca}--\eqref{mcca}.

\ppp{Discrete ICA generalized covariance matrices.} 
Like covariance matrices, generalized covariance matrices of a vector with independent components are diagonal: they satisfy the multilinearity property $\ccal_{D\alpha}(h) = D\,\ccal_{\alpha}(h)D^{\top}$, and are equal to covariance matrices when $h=\zero$.
Therefore, we can expect that the derivations of the diagonal form~\eqref{diagSTdica} of the S-covariance matrices extends to the generalized covariance matrices case.
By analogy with~\eqref{STdica}, we define the \emph{generalized S-covariance matrix} of DICA:
\\[-0.8em]
\begin{equation}\label{GenSdica}
S(t) := \ccal_x(t) - \diag[\ecal_x(t)].
\end{equation}
\\[-1.2em]
To derive the analog of the diagonal form~\eqref{diagSTdica} for $S(t)$, we have to compute all the expectations in~\eqref{gen-expectation} and~\eqref{gen-covariance} for a Poisson random variable $x$ with the parameter $y=D\alpha$. 
To illustrate the intuition, we compute here one of these expectations (see Appendix~\ref{app:epxectations-of-poisson} for further derivations):
\\[-0.6em]
\[
\begin{aligned}
\ebb( xx^{\top} & e^{t^{\top}x} )  = \ebb[ \ebb( xx^{\top} e^{t^{\top} x} \; | \; y ) ]\\
& = \diag[e^t]\, \ebb (yy^{\top} e^{y^{\top}(e^t-\one)} ) \diag[e^t] \\
& = \rbra{\diag[e^t] D}\, \ebb ({\alpha\alpha^{\top} e^{\alpha^{\top} h(t)} }) \rbra{\diag[e^t]D}^{\top}, 
\end{aligned}
\]
\\[-0.4em]
where $h(t) = D^{\top} (e^t-1)$ and $e^t$ denotes an $M$-vector with the $m$-th component equal to $e^{t_m}$.
This gives
\\[-1.5em]
\begin{equation}\label{diagGenSdica}
S(t) = \Rra{\diag[e^t]{D}} \; \ccal_{\alpha}\Rra{h(t)} \; \Rra{\diag[e^t]{D}}^{\top},
\end{equation}
\\[-1.5em]
which is a diagonal form similar (and equivalent for $t=\zero$) to~\eqref{diagSTdica} since the generalized covariance matrix $\ccal_{\alpha}(h)$ of independent sources is diagonal (see~\eqref{gen-covariance-alpha} in Appendix~\ref{app:gen-cums-alpha}). Therefore, the generalized S-covariance matrices, estimated at different processing points $t$, can be used as a substitute of the T-cumulant tensors in the moment matching framework. Interestingly enough, the T-cumulant tensor~\eqref{STdica} can be approximated by the generalized covariance matrix via its directional derivative (see Appendix~\ref{app:dir-derivative-approx}).

\ppp{CCA generalized covariance matrices.} 
For the CCA models~\eqref{ncca}--\eqref{mcca}, straightforward generalizations of the ideas from Section~\ref{sec:cumulants} leads to the following definition of the \emph{generalized CCA S-covariance matrix}:
\\[-1.4em]
\begin{equation}\label{GenSdcca}
S_{12}(t) := \frac{ \ebb ({ x_1x_2^{\top} e^{t^{\top} x} }) }{ \ebb ({ e^{ t^{\top} x } }) } - 
\frac{ \ebb( x_1 e^{t^{\top} x} ) }{ \ebb( e^{t^{\top} x} ) } \frac{ \ebb( x_2^{\top} e^{t^{\top} x} ) }{ \ebb( e^{t^{\top} x} ) },
\end{equation}
\\[-1.4em]
where the vectors $x$ and $t$ are obtained by vertically stacking $x_1$ \& $x_2$ and $t_1$ \& $t_2$ as in~\eqref{stacking}. In the discrete CCA case, $S_{12}(t)$ is essentially the upper-right block of the generalized S-covariance matrix $S(t)$ of DICA and has the form 
\\[-1.4em]
\begin{equation}\label{diagGenSdcca}
S_{12}(t) = \Rra{\diag[e^{t_1}]D_1} \ccal_{\alpha}({ h(t) }) \Rra{ \diag[ e^{t_2}]D_2 }^{\top},
\end{equation}
\\[-1.4em]
where $h(t)=D^{\top}(e^t - \one)$ and the matrix $D$ is obtained by vertically stacking $D_1$ \& $D_2$ by analogy with~\eqref{stacking}.
For non-Gaussian CCA, the diagonal form is
\\[-0.8em]
\begin{equation}\label{diagGenSncca}
S_{12}(t) = D_1 \, \ccal_{\alpha} \rbra{{ h(t) }} \, D_2^{\top},
\end{equation} 
\\[-1.0em]
where $h(t) = D_1^{\top}t_1 + D_2^{\top}t_2$. 
Finally, for mixed CCA,
\\[-0.8em]
\begin{equation}\label{diagGenSmcca}
S_{12}(t) = D_1  \, \ccal_{\alpha} \rbra{{ h(t) }} \, \Rra{ \diag[e^{t_2}]D_2 }^{\top},
\end{equation}
\\[-1.0em]
where $h(t) = D_1^{\top}t_1 + D_2^{\top} (e^{t_2}-\one)$.
Since the generalized covariance matrix of the sources $\ccal_{\alpha}(\cdot)$ is diagonal, expressions~\eqref{diagGenSdcca}--\eqref{diagGenSmcca} have the desired diagonal form (see Appendix~\ref{app:derivations-cca-generalized-covariances} for detailed derivations).

\section{Joint diagonalization algorithms}
\label{sec:jd}

The standard algorithms such as TPM or orthogonal joint diagonalization cannot be used for the estimation of $D_1$ and $D_2$. Indeed, even after whitening, the matrices appearing in the diagonal form~\eqref{diagSdcca}\&\eqref{diagTdcca} or~\eqref{diagGenSdcca}--\eqref{diagGenSmcca} are \emph{not} orthogonal. 
As an alternative, we use Jacobi-like non-orthogonal diagonalization algorithms \citep{FuGao2006,IfeEtAl2009,LucAlb2010}. These algorithms are discussed in this section and in Appendix~\ref{app:nojd}.

The estimation of the factor loading matrices $D_1$ and $D_2$ of the CCA models~\eqref{ncca}--\eqref{mcca} via non-orthogonal joint diagonalization algorithms consists of the following steps: (a) construction of a set of matrices, called \emph{target matrices}, to be jointly diagonalized (using finite sample estimators), (b) a whitening step, (c) a non-orthogonal joint diagonalization step, and (d) the final estimation of the factor loading matrices (Appendix~\ref{app:est-D}).

\ppp{Target matrices.}
There are two ways to construct target matrices: either with the CCA S-matrices~\eqref{Sdcca} and T-cumulants~\eqref{Tdcca} (only DCCA) or the generalized covariance matrices~\eqref{GenSdcca} (D/N/MCCA). These matrices are estimated with finite sample estimators (Appendices~\ref{app:finite-sample-generalized-covariances} \&~\ref{app:finite-sample-cumulants}).

The (computationally efficient) construction of target matrices from S- and T-cumulants was discussed by \citet{PodEtAl2015} and we recall it in Appendix~\ref{app:S-and-T-cumulants}.
Alternatively, the target matrices can be constructed by estimating the generalized S-covariance matrices
at $P+1$ processing points $\zero, t_1,\dots, t_P \in \R^{M_1+M_2}$:
\\[-1.0em]
\begin{equation}\label{jdmatricesGenS}
\{S_{12} = S_{12}(\zero), \;\; S_{12}(t_1), \;\; \dots, \;\; S_{12}(t_P)\},
\end{equation}
\\[-1.3em]
which also have the diagonal form~\eqref{diagGenSdcca}--\eqref{diagGenSmcca}. It is interesting to mention the connection between the T-cumulants and the generalized S-covariance matrices. The T-cumulant can be approximated via the directional derivative of the generalized covariance matrix (see Appendix~\ref{app:dir-derivative-approx}).
However, in general, e.g., $S_{12}(t)$ with $t=[t_1;\zero]$ is not exactly the same as $T_{121}(t_1)$ and the former can be non-zero even when the latter is zero. This is important since order-4 and higher statistics are used with the method of moments when there is a risk that an order-3 statistic is zero like for symmetric sources. 
In general, the use of higher-order statistics increases the sample complexity and makes the resulting expressions quite complicated.
Therefore, replacing the T-cumulants with the generalized S-covariance matrices is potentially beneficial.

\ppp{Whitening.}
The matrices $W_1\in\R^{K\times M_1}$ and $W_2\in\R^{K\times M_2}$ are called \emph{whitening matrices} of $S_{12}$ if
\\[-0.9em]
\begin{equation}\label{whitening}
W_1 S_{12} W_2^{\top} = I_K,
\end{equation}
\\[-1.4em]
where $I_K$ is the $K$-dimensional identity matrix. 
 $W_1$ and $W_2$ are only defined up to multiplication by any invertible matrix $Q\in\R^{K\times K}$, since any pair of matrices $\wt{W}_1=Q W_1$ and $\wt{W}_2 = Q^{-\top} W_2$ also satisfy~\eqref{whitening}. In fact, using higher-order information (i.e. the T-cumulants or the generalized covariances for $t\ne\zero$) allows to solve this ambiguity. 

The whitening matrices can be computed via SVD of $S_{12}$ (see Appendix~\ref{app:compute-whitening}).
When $M_1$ and $M_2$ are too large, one can use a randomized SVD algorithm~\citep[see, e.g.,][]{HalEtAl2011} to avoid the construction of the large matrix $S_{12}$ and to decrease the computational time.

\ppp{Non-orthogonal joint diagonalization (NOJD).}
Let us consider joint diagonalization of the generalized covariance matrices~\eqref{jdmatricesGenS} (the same procedure holds for the S- and T-cumulants~\eqref{jdMatricesS&T}; see Appendix~\ref{app:whiten-t-cumulants}).
Given the whitening matrices $W_1$ and $W_2$, 
the transformation of the generalized covariance matrices~\eqref{jdmatricesGenS} gives $P+1$ matrices
\vspace{-.4em}
\begin{equation}\label{jdmatricesGenSwhitened}
\{W_1S_{12} W_2^{\top}, \; W_1S_{12}(t_p)W_2^{\top}, \quad p=1,\dots,P \},
\vspace{-.4em}
\end{equation}
where each matrix is in $\R^{K\times K}$ and has reduced dimension since $K<M_1,M_2$. In practice, finite sample estimators are used to construct~\eqref{jdmatricesGenS} (see Appendices~\ref{app:finite-sample-generalized-covariances} and~\ref{app:finite-sample-cumulants}).

Due to the diagonal form~\eqref{diagSdcca} and~\eqref{diagGenSdcca}--\eqref{diagGenSmcca}, each matrix in~\eqref{jdmatricesGenS} has the form%
\footnote{
Note that when the diagonal form has terms $\diag[e^t]$, we simply multiply the expression by $\diag[e^{-t}]$.
}
$(W_1D_1) \, \diag(\cdot) \, (W_2D_2)^{\top}$. Both $D_1$ and $D_2$ are (full) $K$-rank matrices and $W_1$ and $W_2$ are $K$-rank by construction. Therefore, the square matrices $V_1 = W_1D_1$ and $V_2 = W_2D_2$ are invertible. From~\eqref{diagSdcca} and~\eqref{whitening}, we get $V_1 \cov(\alpha) V_2^{\top} = I$ and hence $V_2 = \diag[{ \var(\alpha)^{-1} }]V_1^{-1}$ (the covariance matrix of the sources is diagonal and we assume they are non-deterministic, i.e. $\var(\alpha)\ne \zero$). Substituting this into $W_1 S_{12}(t) W_2^{\top}$ and using the diagonal form
~\eqref{diagGenSdcca}--\eqref{diagGenSmcca}, we obtain that the matrices in~\eqref{jdmatricesGenS} have the form $V_1 \diag(\cdot) V_1^{-1}$. Hence, we deal with the problem of the following type:
Given $P$ non-defective (a.k.a.~diagonalizable) matrices
$
\bcal = \cbra{ B_1,\;\dots,\; B_P}$,
where each matrix $B_p \in\R^{K\times K}$, find and invertible matrix $Q\in\R^{K\times K}$ such that 
\\[-.8em]
\begin{equation}\label{nojd}
Q \bcal Q^{-1} = \{ QB_1Q^{-1},\; \dots,\;QB_PQ^{-1} \}
\end{equation}
\\[-1.5em]
are (jointly) as diagonal as possible. This can be seen as a joint non-symmetric eigenvalue problem. 
This problem should not be confused with the classical joint diagonalization problem by congruence (JDC), where $Q^{-1}$ is replaced by $Q^{\top}$, except when $Q$ is an orthogonal matrix \citep{LucAlb2010}. JDC is often used for ICA algorithms or moment matching based algorithms for graphical models when a whitening step is not desirable (see, e.g., \citet{KulEtAl2015} and references therein). However, neither JDC nor the orthogonal diagonalization-type algorithms \citep[such as, e.g., the tensor power method by][]{AnaEtAl2014}  are applicable for the problem~\eqref{nojd}.

\begin{figure*}[!t]
\centering 
\begin{tabular}{ccc}
\includegraphics[width=.3\textwidth,clip=true, trim=5 5 5 5]{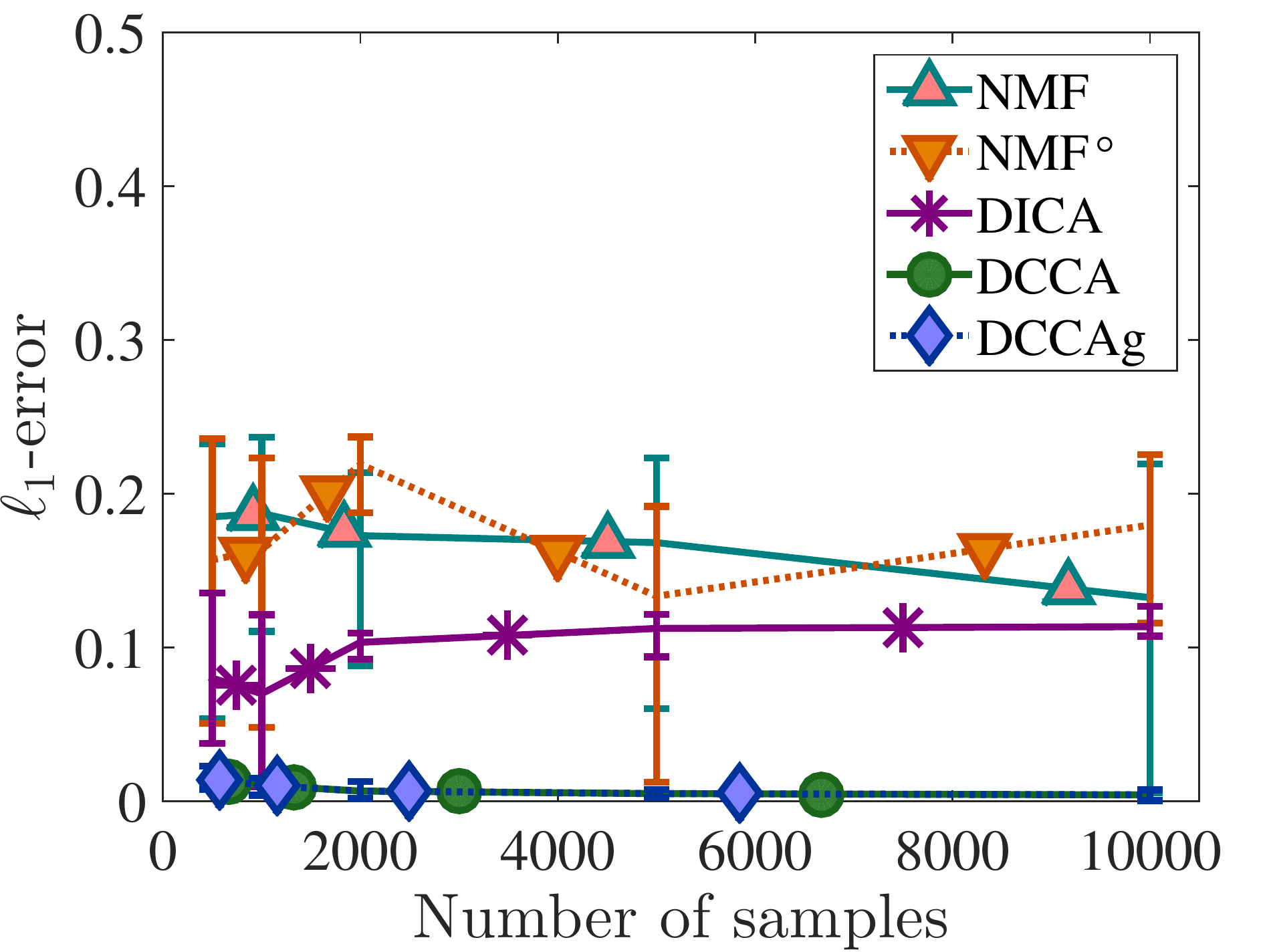} &
\includegraphics[width=.3\textwidth,clip=true, trim=5 5 5 5]{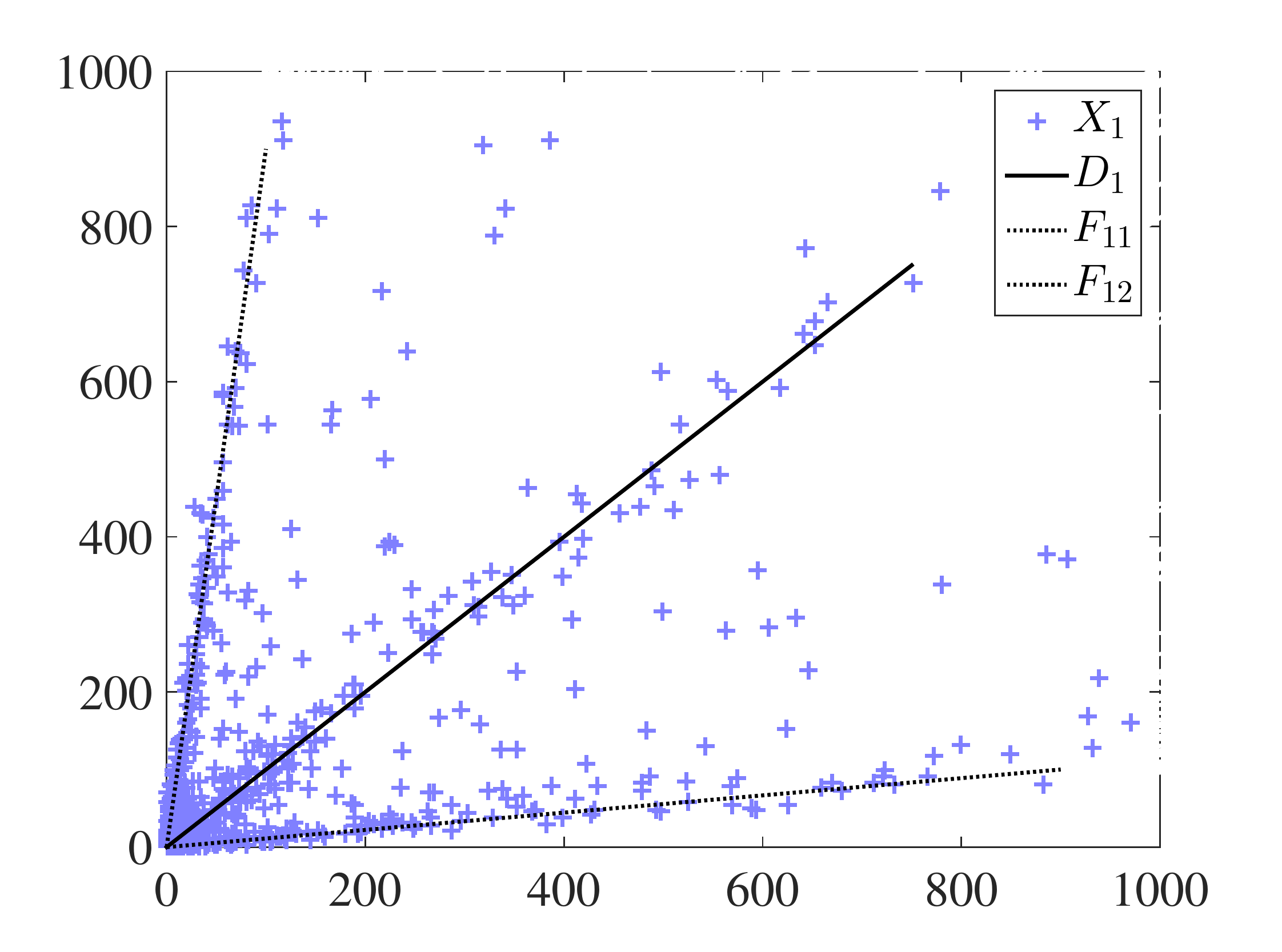} &
\includegraphics[width=.3\textwidth,clip=true, trim=5 5 5 5]{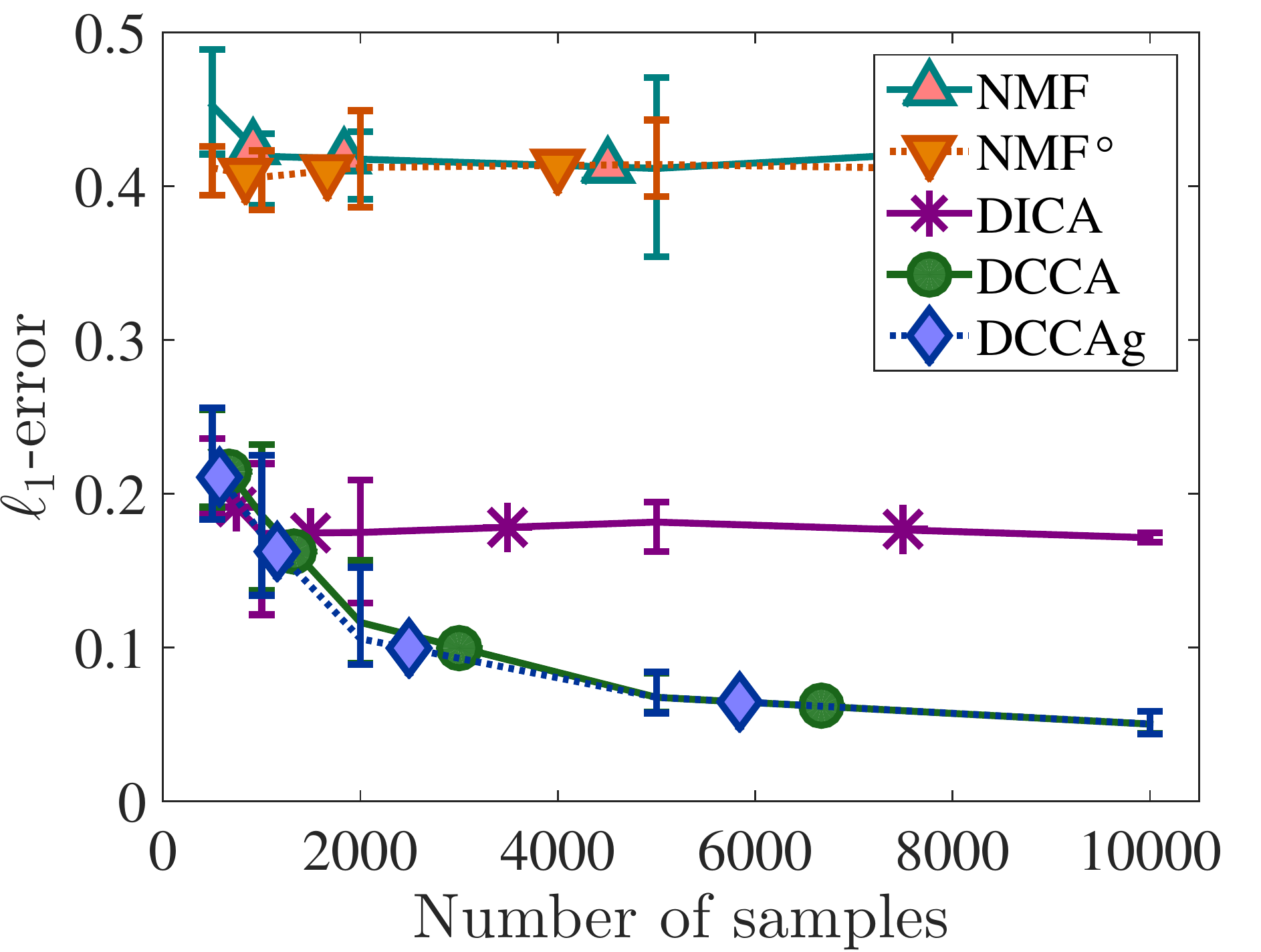}
\end{tabular}
\vspace{-1em}
\caption{Synthetic experiment with discrete data. \textbf{Left (2D example)}: $M_1=M_2=K_1=K_2=2$, $K=1$, $c=c_1=c_2=0.1$, and $L_s = L_n = 100$; \textbf{middle (2D data)}: the $x_1$-observations and factor loading matrices for the 2D example ($F_{1j}$ denotes the $j$-th column of the noise factor matrix $F_1$); \textbf{right (20D example)}: $M_1=M_2=K_1=K_2=20$, $K=10$, $L_s = L_n = 1,000$, $c=0.3$, and $c_1=c_2=0.1$. \vspace{-1.1em}}
\label{fig:dica}
\end{figure*}
\setlength{\textfloatsep}{11pt}
\begin{table*}[!t] 
\centering 
\resizebox{0.85\textwidth}{!}{
\begin{tabular}{cc|cc|cc|cc} 
                 nato &                 otan &                  work &              travail &                 board &           commission &                 nisga &                nisga  \\ 
                kosovo &               kosovo &               workers &        n\'egociations &                 wheat &                 bl\'e &                treaty &          autochtones  \\ 
                forces &           militaires &                strike &         travailleurs &               farmers &         agriculteurs &            aboriginal &              trait\'e  \\ 
              military &               guerre &           legislation &               gr\`eve &                 grain &       administration &             agreement &               accord  \\ 
                   war &        international &                 union &               emploi &             producers &          producteurs &                 right &                droit  \\ 
                troops &                 pays &             agreement &                droit &             amendment &                grain &                  land &              nations  \\ 
               country &           r\'efugi\'es &                labour &             syndicat &                market &              conseil &               reserve &          britannique  \\ 
                 world &            situation &                 right &             services &             directors &                ouest &              national &            indiennes  \\ 
              national &                 paix &              services &               accord &               western &           amendement &               british &                terre  \\ 
                 peace &          yougoslavie &          negotiations &                 voix &              election &              comit\'e &              columbia &             colombie  \\ 
 \end{tabular}
 }
 \vspace{-2.5mm}
\caption{ Factor loadings (a.k.a. topics) extracted from the Hansard collection for $K=20$ with DCCA. 
} 
\label{tab:topics}
\vspace{-1.4em}
\end{table*}

To solve the problem~\eqref{nojd}, we use the Jacobi-like non-orthogonal joint diagonalization (NOJD) algorithms  \citep[e.g.,][]{FuGao2006,IfeEtAl2009,LucAlb2010}. 
These algorithms are an extension of the orthogonal joint diagonalization algorithms based on Jacobi (=Givens) rotations \citep{GolVan1996,BunEtAl1993,CarSou1996}.
Due to the space constraint, the description of the NOJD algorithms is moved to Appendix~\ref{app:nojd}.
Although these algorithms are quite stable in practice, we are not aware of any theoretical guarantees about their convergence or stability to perturbation.

\ppp{Spectral algorithm.} By analogy with the orthogonal case \citep{Car1989,AnaEtAl2012}, we can easily extend the idea of the spectral algorithm to the non-orthogonal one. Indeed, it amounts to performing whitening as before and constructing only one matrix with the diagonal structure, e.g., $B=W_1 S_{12}(t) W_2^{\top}$ for some $t$. Then, the matrix $Q$ is obtained as the matrix of the eigenvectors of $B$. The vector $t$ can be, e.g., chosen as $t=Wu$, where $W=[W_1; \,W_2]$ and $u\in\R^K$ is a vector sampled uniformly at random.

This spectral algorithm and the NOJD algorithms are closely connected. In particular, when $B$ has real eigenvectors, the spectral algorithm is equivalent to NOJD of $B$. Indeed, in such case, NOJD boils down to an algorithm for a non-symmetric eigenproblem \citep{Ebe1962,Ruh1968}. 
In practice, however, due to the presence of noise and finite sample errors, $B$ may have complex eigenvectors. In such case, the spectral algorithm is different from NOJD.
Importantly, the joint diagonalization type algorithms are known to be more stable in practice \citep[see, e.g.,][]{BacJor2002,PodEtAl2015}.

While deriving precise theoretical guarantees is beyond the scope of this paper, the techniques outlined by \citet{AnaEtAl2012} for the spectral algorithm for latent Dirichlet Allocation can potentially be extended. The main difference is obtaining the analogue of the SVD accuracy \citep[Lemma C.3,][]{AnaEtAl2013} for the eigen decomposition. This kind of analysis can potentially be extended with the techniques outlined in \citep[Chapter 4,][]{SteSun1990}.
Nevertheless, with appropriate parametric assumptions on the sources, we expect that the above described extension of the spectral algorithm should lead to similar guarantee as the spectral algorithm of \citet{AnaEtAl2012}.  

See Appendix~\ref{app:implementation-details} for some important implementation details, including the choice of the processing points.

\section{Experiments}\label{sec:experiments}

\ppp{Synthetic data.} 
We sample synthetic data to have ground truth information for comparison.
We sample from linear DCCA which extends linear CCA~\eqref{linear-cca} such that each view is $x_j \sim \poi (D_j \alpha + F_j \beta_j )$. The sources $\alpha \sim \gam(c,b)$ and the noise sources $\beta_j \sim \gam(c_j,b_j)$, for $j=1,2$, are sampled from the gamma distribution (where $b$ is the rate parameter). Let $s_j \sim \poi(D_j \alpha)$ be the part of the sample due to the sources and $n_j \sim \poi(F_j\beta_j)$ be the part of the sample due to the noise (i.e., $x_j = s_j + n_j$). Then we define the expected sample length due to the sources and noise, respectively, as $L_{js} := \ebb[\sum\mathop{}_{m} s_{jm}]$ and $L_{jn} := \ebb[ \sum\mathop{}_{m} n_{jm}]$. For sampling, the target values $L_s = L_{1s} = L_{2s}$ and $L_n = L_{1n} = L_{2n}$ are fixed and the parameters $b$ and $b_j$ are accordingly set to ensure these values: $b= Kc / L_s$ and $b_j = K_j c_j / L_n$ (see Appendix B.2 of \citet{PodEtAl2015}). For the larger dimensional example (Fig.~\ref{fig:dica}, right), each column of the matrices $D_j$ and $F_j$, for $j=1,2$, is sampled from the symmetric Dirichlet distribution with the concentration parameter equal to $0.5$. For the smaller 2D example (Fig.~\ref{fig:dica}, left), they are fixed: $D_1=D_2$ with $[D_1]_1=[D_1]_2=0.5$ and $F_1=F_2$ with $[F_1]_{11}=[F_1]_{22}=0.9$ and $[F_1]_{12} = [F_1]_{21}=0.1$.
For each experiment, $D_j$ and $F_j$, for $j=1,2$, are sampled once and, then, the $x$-observations are sampled for different sample sizes $N = \cbra{ 500, 1,000, 2,000, 5,000, 10,000}$, 5 times for each $N$.

\ppp{Metric.} The evaluation is performed on a matrix $D$ obtained by stacking $D_1$ and $D_2$ vertically (see also the comment after Thm.~\ref{thm:identif}). As in~\citet{PodEtAl2015}, we use as evaluation metric the normalized $\ell_1$-error between a recovered matrix $\wh{D}$ and the true matrix $D$ with the best permutation of columns $\mathrm{err}_1(\wh{D},D):=\min_{\pi\in\mathrm{PERM}} \frac{1}{2K} \sum_k \|{ \wh{d}_{\pi_k} - d_k }\|_1 \in [0,1]$. The minimization is over the possible permutations $\pi\in\mathrm{PERM}$ of the columns of $\wh{D}$ and can be efficiently obtained with the Hungarian algorithm for bipartite matching. The (normalized) $\ell_1$-error takes the values in $[0,\,1]$ and smaller values of this error indicate better performance of an algorithm.

\ppp{Algorithms.} We compare DCCA (implementation with the S- and T-cumulants) and DCCAg (implementation with the generalized S-covariance matrices and the processing points initialized as described in Appendix~\ref{app:processing-projection-points}) to DICA and the non-negative matrix factorization (NMF) algorithm with multiplicative updates for divergence \citep{LeeSeu2000}. To run DICA or NMF, we use the stacking trick~\eqref{stacking}. DCCA is set to estimate $K$ components. DICA is set to estimate either $K_0 = K+K_1+K_2$ or $M=M_1+M_2$ components (whichever is the smallest, since DICA cannot work in the over-complete case).  NMF is always set to estimate $K_0$ components. For the evaluation of DICA/NMF, the $K$ columns  with the smallest $\ell_1$-error are chosen. NMF$^{\circ}$ stands for NMF initialized with a matrix $D$ of the form~\eqref{stacking} with induced zeros; otherwise NMF is initialized with (uniformly) random non-negative matrices. The running times are discussed in Appendix~\ref{app:running-time}.

\ppp{Synthetic experiment.} We first perform an experiment with discrete synthetic data in 2D (Fig.~\ref{fig:dica}) and then repeat the same experiment when the size of the problem is 10 times larger. In practice, we observed that for $K_0<M$ all models work approximately equally well, except for NMF which breaks down in high dimensions. In the over-complete case as in Fig.~\ref{fig:dica}, DCCA works better. 
A continuous analogue of this experiment is presented in Appendix~\ref{sec:exp-continuous}.

\ppp{Real data (translation).}
Following \citet{VinEtAl2003}, we illustrate the performance of DCCA by extracting bilingual topics from the Hansard collection \citep{VinGir2002} with aligned English and French proceedings of the 36-th Canadian Parliament. 
In Table~\ref{tab:topics}, we present some of the topics extracted after running DCCA with $K=20$ (see all the details in Appendices~\ref{app:translation} and~\ref{app:translation-preprocess}). The (Matlab/C++) code for reproducing the experiments of this paper is available at \href{https://github.com/anastasia-podosinnikova/cca}{https://github.com/anastasia-podosinnikova/cca}.

\subsection*{Conclusion} We have proposed the first identifiable versions of CCA, together with moment matching algorithms which allow the identification of the loading matrices in a semi-parametric framework, where no assumptions are made regarding the distribution of the source or the noise. We also introduce new sets of moments (our generalized covariance matrices), which could prove useful in other settings.

\section*{Acknowledgements} 
This work was partially supported by the MSR-Inria Joint Center.

\bibliography{lit}
\bibliographystyle{icml2016}

\cleardoublepage
\newpage
\section{Appendix}
\appendix
\renewcommand{\thesubsection}{\Alph{subsection}}

The appendix is organized as follows. 
\begin{mi}
\item
In Appendix~\ref{app:notation}, we summarize our notation.
\item
In Appendix~\ref{app:identifiability}, we present the proof of Theorem~\ref{thm:identif} stating the \emph{identifiability of the CCA models~\eqref{ncca}--\eqref{mcca}}.
\item
In Appendix~\ref{app:generalized-E-cov}, we provide some details for the \emph{generalized covariance matrices}: the form of the generalized covariance matrices of independent variables (Appendix~\ref{app:gen-cums-alpha}), the derivations of the diagonal form of the generalized covariance matrices of discrete ICA (Appendix~\ref{app:derivations-dica-generalized-covariances}), the derivations of the diagonal form of the generalized covariance matrices of the CCA models~\eqref{ncca}--\eqref{mcca} (Appendix~\ref{app:derivations-cca-generalized-covariances}), and approximation of the T-cumulants with the generalized covariance matrix (Appendix~\ref{app:dir-derivative-approx}).
\item
In Appendix~\ref{app:finite-sample-estimators}, we provide expressions for natural \emph{finite sample estimators of the generalized covariance matrices} and the T-cumulant tensors for the considered CCA models.
\item
In Appendix~\ref{app:implementation-details}, we discuss some rather technical \emph{implementation details}: computation of whitening matrices (Appendix~\ref{app:compute-whitening}), 
selection of the projection vectors for the T-cumulants and the processing points for the generalized covariance matrices (Appendix~\ref{app:processing-projection-points}), and the final estimation of the factor loading matrices (Appendix~\ref{app:est-D}).
\item
In Appendix~\ref{app:nojd}, we describe the \emph{non-orthogonal joint diagonalization algorithms} used in this paper.
\item In Appendix~\ref{app:supplementary-experiments}, we present \emph{some supplementary experiments}: a continuous analog of the synthetic experiment from Section~\ref{sec:experiments} (Appendix~\ref{sec:exp-continuous}), an experiment to analyze the sensitivity of the DCCA algorithm with the generalized S-covariance matrices to the choice of the processing points (Appendix~\ref{app:gdcca-vs-delta}), and a detailed description of the experiment with the real data from Section~\ref{sec:experiments} (Appendices~\ref{app:translation} and~\ref{app:translation-preprocess}).
\end{mi}

\subsection{Notation}
\label{app:notation}

\subsubsection{Notation summary}
\label{app:notation-summary}
The vector $\alpha\in\R^K$ refers to the latent sources. Unless otherwise specified, the components $\alpha_1,\dots,\alpha_K$ of the vector $\alpha$ are mutually independent. For a linear single-view model, $x=D\alpha$, the vector $x\in\R^M$ denotes the observation vector (sensors or documents), where $M$ is, respectively, the number of sensors or the vocabulary size. For the two-view model, $x$, $M$, and $D$ take the indices $1$ and $2$.

\subsubsection{Naming convention}
\label{app:naming_convention}
A number of models have the linear form $x=D\alpha$.
Depending on the context, the matrix $D$ is called differently:   topic matrix%
\footnote{ Note that \citet{PodEtAl2015} show that DICA is closely connected (and under some conditions is equivalent) to latent Dirichlet allocation \citep{BleEtAl2003}. 
} 
in the topic learning context,
factor loading or projection matrix in the FA and/or PPCA context,  mixing matrix in the ICA context, or  dictionary in the dictionary learning context. 

Our linear multi-view models, $x_1 = D_1\alpha$ and $x_2 = D_2\alpha$, are closely related the linear models mentioned above.
For example, due to the close relation of DCCA and DICA, the former is closely related to the multi-view topic models \citepsup[see, e.g.,][]{BleJor2003}.
In this paper, we refer to $D_1$ and $D_2$ as the  \emph{factor loading matrices}, although depending on contex any other name can be used.

\subsection{Identifiability}
\label{app:identifiability}
In this section, we prove that the factor loading matrices $D_1$ and $D_2$ of the non-Gaussian CCA~\eqref{ncca}, discrete CCA~\eqref{dcca}, and mixed CCA~\eqref{mcca} models are identifiable up to permutation and scaling if at most one source $\alpha_k$ is Gaussian.
We provide a complete proof for the non-Gaussian CCA case and show that the other two cases can be proved by analogy.

\subsubsection{Identifiability of non-Gaussian CCA~\eqref{ncca}}
The proof uses the notion of the second characteristic function (SCF) of a random variable $x\in\R^M$:
$$
\phi_x(t) = \log \ebb ( e^{i t^{\top} x} ),
$$
for all $t\in\R^M$.
The SCF completely defines the probability distribution of $x$ \citepsup[see, e.g.,][]{JacPro2004}.
Important difference between the SCF and the cumulant generating function~\eqref{cgf} is that the former always exists.

The following property of the SCF is of central importance for the proof: if two random variables, $z_1$ and $z_2$, are independent, then $\phi_{A_1z_1+A_2z_2}(t) = \phi_{z_1}(A_1^{\top}t) + \phi_{z_2}(A_2^{\top}t)$, where $A_1$ and $A_2$ are any  matrices of compatible sizes.

We can now use our CCA model to derive an expression of $\phi_x(t)$.
Indeed, defining a vector $x$ by stacking the vectors $x_1$ and $x_2$, the SCF of $x$ for any $t=[t_1;\;t_2]$, takes the form
$$
\begin{aligned}
\phi_x&(t) = \log \ebb ( e^{it_1^{\top} x_1 + it_2^{\top} x_2} ) \\
&\stackrel{(a)}{=} \log \ebb ( e^{ i\alpha^{\top}(D_1^{\top} t_1 + D_2^{\top} t_2) + i \varepsilon_1^{\top} t_1 + i \varepsilon_2^{\top} t_2 } ) \\
&\stackrel{(b)}{=} \log \ebb (e^{ i \alpha^{\top} (D_1^{\top} t_1 + D_2^{\top} t_2) })  \\
& \hspace*{2cm} + \log \ebb (e^{i\varepsilon_1^{\top} t_1}) + \log \ebb (e^{i\varepsilon_2^{\top} t_2}) \\
&= \phi_{\alpha} (D_1^{\top}t_1 + D_2^{\top} t_2) + \phi_{\varepsilon_1}(t_1) + \phi_{\varepsilon_2}(t_2),
\end{aligned}
$$
where in $(a)$ we substituted the definition~\eqref{ncca} of~$x_1$ and~$x_2$ and in $(b)$ we used the independence $\alpha \ci \varepsilon_1 \ci \varepsilon_2$. Therefore, the blockwise mixed derivatives of $\phi_x$ are equal to
\begin{equation}\label{temp:hessianSCF}
 \partial_1 \partial_2 \phi_{x}(t) = D_1 \phi''_{\alpha} (D_1^{\top} t_1 + D_2^{\top} t_2) D_2^{\top},
\end{equation}
where $\partial_1 \partial_2 \phi_{x}(t) := \nabla_{t_1}\nabla_{t_2} \phi_{x}(h(t_1,t_2))\in\R^{M_1\times M_2}$ and $\phi''_{\alpha}(u) :=\nabla_u^2\phi_{\alpha}(u)$, does not depend on the noise vectors $\varepsilon_1$ and $\varepsilon_2$. 

For simplicity, we first prove the identifiability result when all components of the common sources are non-Gaussian.
The high level idea of the proof is as follows. We assume two different representations of $x_1$ and $x_2$ and using~\eqref{temp:hessianSCF} and the independence of the components of $\alpha$ and the noises, we first show that the two potential dictionaries are related by an orthogonal matrix (and not any invertible matrix), and then show that this implies that the two potential sets of independent components are (orthogonal) linear combinations of each other, which, for non-Gaussian components which are not reduced to point masses, imposes that this orthogonal transformation is the combination of a permutation matrix and marginal scaling---a standard result from the ICA literature \citep[Theorem 11]{Com1994}.

Let us then assume that two equivalent representations of non-Gaussian CCA exist:
\begin{equation}\label{temp:tworepres}
\begin{aligned}
x_1 &= D_1 \alpha + \varepsilon_1 = E_1 \beta + \eta_1,\\
x_2 &= D_2 \alpha + \varepsilon_2 = E_2 \beta + \eta_2,
\end{aligned}
\end{equation}
where the other sources $\beta = (\beta_1, \ldots, \beta_K)$ are also assumed mutually independent and non-degenerate.
As a standard practice in the ICA literature and without loss of generality as the sources have non-degenerate components, one can assume that the sources have unit variances, i.e. $\cov(\alpha,\alpha)=I$ and $\cov(\beta, \beta)=I$, by respectively rescaling the columns of the factor loading matrices. Under this assumption, the two expressions of the cross-covariance matrix are
\begin{equation} \label{eq:D1D2}
\cov(x_1,x_2) = D_1D_2^{\top} = E_1 E_2^{\top},
\end{equation}
which, given that $D_1$, $D_2$ have full rank, 
implies that\footnote{The fact that $D_1$, $D_2$ have full rank and that $E_1$, $E_2$ have $K$ columns, combined with $\eqref{eq:D1D2}$, implies that $E_1$, $E_2$ have also full rank.}
\begin{equation}\label{temp:E1E2}
E_1 = D_1 Q, \quad E_2 = D_2 Q^{-\top},
\end{equation}
where $Q\in\R^{K\times K}$ is some invertible matrix. Substituting the representations~\eqref{temp:tworepres} into the blockwise mixed derivatives of the SCF~\eqref{temp:hessianSCF} and using the expressions~\eqref{temp:E1E2} give
$$
\begin{aligned}
D_1  \phi''_{\alpha} & (D_1^{\top}t_1 + D_2^{\top}t_2) D_2^{\top} \\
&= D_1 Q \phi''_{\beta} (Q^{\top} D_1^{\top} t_1 + Q^{-1}D_2^{\top}t_2) Q^{-1}D_2^{\top}, 
\end{aligned}
$$
for all $t_1\in\R^{M_1}$ and $t_2\in\R^{M_2}$. Since the matrices $D_1$ and $D_2$ have full rank, this can be rewritten as
$$
\begin{aligned}
 \phi''_{\alpha} & (D_1^{\top}t_1 + D_2^{\top}t_2) \\
&= Q \phi''_{\beta} (Q^{\top} D_1^{\top} t_1 + Q^{-1}D_2^{\top}t_2) Q^{-1},
\end{aligned}
$$
which holds for all $t_1\in\R^{M_1}$ and $t_2\in\R^{M_2}$. Moreover, still since $D_1$ and $D_2$ have full rank, we have, for any $u_1, u_2 \in \R^K$ the existence of   $t_1\in\R^{M_1}$ and $t_2\in\R^{M_2}$, such that $u_1 = D_1^\top t_1 $ and $u_2 = D_2^\top t_2$, that is,
\begin{equation}\label{temp:hessians-realtion}
 \phi''_{\alpha} (u_1 + u_2) = Q \phi''_{\beta} (Q^{\top} u_1 + Q^{-1}u_2) Q^{-1},
\end{equation}
for all $u_1,u_2\in\R^K$.

We will now prove two facts:
\begin{enumerate}
\item[(F1)] For any vector $v \in \R^K$, then $\phi''_{\beta} ((Q^{\top} Q -I) v) = - I$, which will imply that $QQ^{\top} = I$ because of the non-Gaussian assumptions.
\item[(F2)] If $QQ^{\top} = I$, then $\phi''_{\alpha}(u) = \phi''_{Q\beta}(u)$ for any $u\in\R^{K}$, which will imply that $Q$ is the composition of a permutation and a scaling. This will end the proof.
\end{enumerate}

\emph{Proof of fact (F1).} By letting $u_1 = Qv$ and $u_2 = - Qv$, we get:
\begin{equation}\label{temp:1}
\phi''_{\alpha} (0) = Q \phi''_{\beta} ( (Q^{\top}Q-I)v) Q^{-1},
\end{equation}
 Since%
\footnote{
Note that $\nabla_u^2\phi_{\alpha}(u) = - \frac{\ebb ({ \alpha \alpha^{\top} e^{iu^{\top}\alpha}})}{\ebb (e^{i u^{\top} \alpha})} + \ecal_{\alpha}(u)\ecal_{\alpha}(u)^{\top}$, where $\ecal_{\alpha}(u)=\frac{\ebb ({ \alpha e^{iu^{\top}\alpha}})}{\ebb (e^{i u^{\top} \alpha})}$.
}
$\phi''_{\alpha}(0)=-\cov(\alpha) = -I$, one gets
$$
\phi''_{\beta} ((Q^{\top}  Q- I)v) = -I,
$$
for any $v\in\R^K$.

Using the property that $\phi''_{A^\top \beta}(v) = A^\top \phi''_\beta(A v) A$ for any matrix $A$, and in particular with $A = Q^{\top} Q - I$, we have that $\phi''_{A^\top \beta}(v) = -A^\top A$, i.e. is constant.

If the second derivative of a function is constant, the function is quadratic. Therefore, $\phi_{A^\top \beta}(\cdot)$ is a quadratic function. Since the SCF completely defines the distribution of its variable (see,e.g., \citetsup{JacPro2004}), $A^\top \beta$ must be Gaussian (the SCF of a Gaussian random variable is a quadratic function). Given Lemma 9 from \citet{Com1994} (i.e., Cramer's lemma: a linear combination of non-Gaussian random variables cannot be Gaussian unless the coefficients are all zero), this implies that $A=0$, and hence $Q^{\top} Q = I$, i.e., $Q$ is an orthogonal matrix.

\emph{Proof of fact (F2)}. Plugging $Q^{\top}=Q^{-1}$ into~\eqref{temp:hessians-realtion}, with $u_1=0$ and $u_2=u$, gives
\begin{equation}\label{temp:2}
\phi''_{\alpha}(u) = Q \phi''_{\beta}(Q^{\top} u) Q^{\top} = \phi''_{Q\beta}(u),
\end{equation}
for any $u\in\R^{K}$.
By integrating both sides of~\eqref{temp:2} and using $\phi_\alpha(0) = \phi_{Q\beta}(0) = 0$, we get that $\phi_\alpha(u) = \phi_{Q\beta}(u) + i \gamma^\top u$ for all $u\in\R^{K}$ for some constant vector $\gamma$. 
Using again that the SCF completely defines the distribution, it follows that $\alpha-\gamma$ and $Q\beta$ have the same distribution. Since both $\alpha$ and $\beta$ have independent components, this is only possible when $Q=\Lambda P$, where $P$ is a permutation matrix and $\Lambda$  is some diagonal matrix \citep[Theorem 11]{Com1994}.

\subsubsection{Case of a single Gaussian source}
Without loss of generality, we assume that the potential Gaussian source is the first one for $\alpha$ and $\beta$.
The first change is in the proof of fact (F1). We use the same argument up to the point where we conclude that $A^\top \beta$ is a Gaussian vector. As only $\beta_1$ can be Gaussian, Cramer's lemma implies that only the first row of $A$ can have non-zero components, that is $A = Q^\top Q  - I = e_1 f^\top$, where $e_1$ is the first basis vector and $f$ any vector. Since $Q^\top Q$ is symmetric, we must have 
$$
Q^\top Q  = I + a e_1 e_1^{\top},
$$
where  $a$ is a constant scalar different than $-1$ as $Q^\top Q $ is invertible. This implies that
$Q^\top Q$ is an invertible diagonal matrix $\Lambda$, and hence
$Q \Lambda^{-1/2}$ is an orthogonal matrix, which in turn implies that $Q^{-1} = \Lambda^{-1} Q^\top$.

Plugging this into~\eqref{temp:hessians-realtion} gives, for any $u_1$ and $u_2$:
$$
 \phi''_{\alpha} (u_1 + u_2) = Q \phi''_{\beta} (Q^{\top} u_1 + \Lambda^{-1} Q^{\top}u_2) \Lambda^{-1} Q^{\top}.
$$
Given that diagonal matrices commute and that $\phi''_{\beta}$ is diagonal for independent sources (see Appendix~\ref{app:gen-cums-alpha}), this leads to
$$
 \phi''_{\alpha} (u_1 + u_2) = Q \Lambda^{-1/2} \phi''_{\beta} (Q^{\top} u_1 + \Lambda^{-1} Q^{\top}u_2) \Lambda^{-1/2} Q^{\top}.
$$
For any given $v \in \R^K$, we are looking for $u_1$ and $u_2$ such that 
$Q^{\top} u_1 + \Lambda^{-1} Q^{\top}u_2 = \Lambda^{-1/2} Q^\top  v$ and $u_1+u_2=v$, which is always possible by setting
$Q^\top u_2 = (\Lambda^{-1/2} + I)^{-1} Q^\top v$ and $Q^\top u_1 = Q^\top v - Q^\top u_2$ by using the special structure of $\Lambda$. Thus, for any $v$,
$$
 \phi''_{\alpha} (v) = Q \Lambda^{-1/2} \phi''_{\beta} (\Lambda^{-1/2} Q^\top  v) \Lambda^{-1/2} Q^{\top}
 =  \phi''_{Q \Lambda^{-1/2} \beta} ( v).
$$
Integrating as previously, this implies that the characteristic function of $\alpha$ and $Q \Lambda^{-1/2} \beta $ differ only by a linear function $i \gamma^\top v$, and thus, that $\alpha - \gamma$ and $Q \Lambda^{-1/2} \beta$ have the same distribution. This in turn, from \citet[Theorem 11]{Com1994}, implies that $Q \Lambda^{-1/2} $ is a product of a scaling and a permutation, which ends the proof.

\subsubsection{Identifiability of discrete CCA~\eqref{dcca} and mixed CCA~\eqref{mcca}}
Given the discrete CCA model, the SCF $\phi_x(t)$ takes the form
$$
\begin{aligned}
\phi_x(t) &= \phi_{\alpha} (D_1^{\top}(e^{it_1}-\one)+D_2^{\top}(e^{it_2}-\one))\\
&+ \phi_{\varepsilon_1}(e^{it_1}-\one) + \phi_{\varepsilon_2}(e^{it_2}-\one),
\end{aligned}
$$
where $e^{it_j}$, for $j=1,2$,  denotes a vector with the $m$-th element equal to $e^{i[t_j]_m}$, and we used the arguments analogous with the non-Gaussian case. The rest of the proof extends with a correction that sometimes one has to replace $D_j$ with $\diag[e^{it_j}]D_j$ and that $u_j =D_j^{\top}(e^{it_j}-\one)$ for $j=1,2$. For the mixed CCA case, only the part related to $x_2$ and $D_2$ changes in the same way as for the discrete CCA case.

\subsection{The generalized expectation and covariance matrix}
\label{app:generalized-E-cov}

\subsubsection{The generalized expectation and covariance matrix of the sources}
\label{app:gen-cums-alpha}
Note that some properties of the generalized expectation and covariance matrix, defined in~\eqref{gen-expectation} and~\eqref{gen-covariance}, and their natural finite sample estimators are analyzed by \citet{SlaYer2012b}. Note also that we find the name ``generalized covariance matrix'' to be more meaningful than ``charrelation'' matrix as was proposed by previous authors~\citep[see, e.g.][]{SlaYer2012,SlaYer2012b}.

The sources $\alpha=(\alpha_1,\dots,\alpha_K)$ are mutually independent. Therefore, for some $h\in\R^K$, their CGF~\eqref{cgf} $K_{\alpha}(h) = \log\ebb({ e^{\alpha^{\top} h} })$ takes the form
$$
K_{\alpha}(h) = \sumk \log\sbra{\ebb ( e^{\alpha_k h_k} ) }.
$$ 
Therefore, the $k$-th element of the generalized expectation~\eqref{gen-expectation} of $\alpha$ is (separable in $\alpha_k$)
\begin{equation}\label{gen-expectation-alpha}
\sbra{\ecal_{\alpha}(h)}_k = \frac{\ebb ( \alpha_k e^{\alpha_k h_k} )}{\ebb ( e^{\alpha_k h_k} ) }
\end{equation}
and the generalized covariance~\eqref{gen-covariance} of $\alpha$ is diagonal due to the separability and its $k$-th diagonal element is
\begin{equation}\label{gen-covariance-alpha}
\sbra{\ccal_{\alpha}(h)}_{kk} = \frac{ \ebb (\alpha^2_k e^{\alpha_k h_k}) }{\ebb ( e^{\alpha_k h_k} ) } - \sbra{ \ecal_{\alpha} (h)}_k^2.
\end{equation}

\subsubsection{Some expectations of a Poisson random variable}
\label{app:epxectations-of-poisson}

Let $x\in\R^M$ be a multivariate Poisson random variable with mean $y\in\R_+^{M}$. Then, for some $t\in\R^M$,
$$
\begin{aligned}
\ebb ({ e^{t^{\top}x } }) &= e^{ y^{\top}(e^t - 1) }, \\
\ebb ({ x_m e^{t^{\top} x} }) &= y_m e^{t_m} e^{ y^{\top}(e^t - 1) }, \\
\ebb ({ x_m^2 e^{t^{\top} x } }) &= \sbra{y_m e^{t_m}+1}y_m e^{t_m} e^{ y^{\top}(e^t - 1) }, \\
\ebb ({ x_{m}x_{m'} e^{t^{\top} x} }) &= y_{m} e^{t_{m}} y_{m'} e^{t_{m'}} e^{ y^{\top}(e^t - 1) }, \quad m\ne m',
\end{aligned}
$$
where $e^t$ denotes an $M$-vector with the $m$-th element equal to $e^{t_m}$.

\subsubsection{The generalized expectation and covariance matrix of discrete ICA}
\label{app:derivations-dica-generalized-covariances}
In this section, we use the expectations of a Poisson random variable presented in Appendix~\ref{app:epxectations-of-poisson}.
 
Given the discrete ICA model~\eqref{dica}, the generalized expectation~\eqref{gen-expectation} of $x\in\R^M$ takes the form
$$
\begin{aligned}
\ecal_x(t) &= \frac{ \ebb (x e^{t^{\top} x}) }{ \ebb (e^{t^{\top} x}) } 
 = \frac{ \ebb \sbra{ \ebb ( x e^{t^{\top} x} | \alpha )} }{ \ebb \sbra{ \ebb (e^{t^{\top} x} | \alpha )} } \\
& = \diag[e^t] D \frac{ \ebb (\alpha e^{\alpha^{\top} h(t) }) }{ \ebb( e^{\alpha^{\top} h(t) }) } \\
& = \diag[e^t] D \ecal_{\alpha} (h(t)),
\end{aligned}
$$
where $t\in\R^M$ is a parameter, $h(t)=D^{\top} (e^t - 1)$, and~$e^t$ denotes an~$M$-vector with the~$m$-th element equal to~$e^{t_m}$. 
Note that in the last equation we used the definition~\eqref{gen-expectation} of the generalized expectation $\ecal_{\alpha} (\cdot)$.

Further, the generalized covariance~\eqref{gen-covariance} of $x$ takes the form
$$
\begin{aligned}
\ccal_x(t) & = \frac{\ebb (xx^{\top} e^{t^{\top}x}) }{\ebb ( e^{t^{\top}x} ) } - \ecal_x(t) \ecal_x(t)^{\top} \\
& = \frac{\ebb \sbra{ \ebb (xx^{\top} e^{t^{\top}x} | \alpha)} }{\ebb\sbra{ \ebb ( e^{t^{\top}x} | \alpha) }  }- \ecal_x(t) \ecal_x(t)^{\top}.
\end{aligned}
$$
Plugging into this expression the expression for~$\ecal_x(t)$ and 
$$
\begin{aligned}
\ebb (xx^{\top} e^{t^{\top}x} | \alpha) &= \diag[e^t]D\ebb ({\alpha\alpha^{\top}e^{\alpha^{\top}h(t)}}) D^{\top} \diag[e^t] \\
&+ \diag[e^t]\diag\sbra{D\ebb ({\alpha e^{\alpha^{\top}h(t)}}) }
\end{aligned}
$$
we get
$$
\ccal_x(t)  = \diag[\ecal_x(t)] + \diag[e^t]{D} \ccal_{\alpha} (h(t)) {D}^{\top}\diag[e^t],
$$
where we used the definition~\eqref{gen-covariance} of the generalized covariance of $\alpha$.

\subsubsection{The generalized CCA S-covariance matrix}
\label{app:derivations-cca-generalized-covariances}
In this section we sketch the derivation of the diagonal form~\eqref{diagGenSmcca} of the generalized S-covariance matrix of mixed CCA~\eqref{mcca}. Expressions~\eqref{diagGenSdcca} and~\eqref{diagGenSncca} can be obtained in a similar way.

Denoting $x=[x_1;\;x_2]$ and $t=[t_1;\;t_2]$ (i.e. stacking the vectors as in~\eqref{stacking}), the CGF~\eqref{cgf} of mixed CCA~\eqref{mcca} can be written as
$$
\begin{aligned}
&K_x(t)  = \log \ebb ({ e^{t_1^{\top} x_1 + t_2^{\top} x_2 } }) \\
&= \log \ebb\sbra{ \ebb({e^{t_1^{\top} x_1 + t_2^{\top} x_2} |\, \alpha, \varepsilon_1, \varepsilon_2}) }\\
& \stackrel{(a)}{=} \log \ebb\sbra{ \ebb({e^{t_1^{\top}x_1} |\, \alpha, \varepsilon_1}) \ebb({e^{t_2^{\top}x_2}|\,\alpha,\varepsilon_2}) } \\
& \stackrel{(b)}{=} \log \ebb\Rra{ e^{t_1^{\top}(D_1\alpha + \varepsilon_1)}e^{(D_2\alpha + \varepsilon_2)^{\top}(e^{t_2}-1)} } \\
& \stackrel{(c)}{=} \log\ebb\Rra{ e^{\alpha^{\top} h(t)} }  
 + \log \ebb \Rra{ e^{\varepsilon_2^{\top}(e^{t_2}-1)} } + \log\ebb ({ e^{t_1^{\top} \varepsilon_1 } }),
\end{aligned}
$$
where $h(t)=({D_1^{\top}t_1 + D_2^{\top}(e^{t_2} - 1})$, in $(a)$ we used the conditional independence of $x_1$ and $x_2$, in $(b)$ we used the first expression from Appendix~\ref{app:epxectations-of-poisson}, and in $(c)$ we used the independence assumption~\eqref{indep}.

The generalized CCA S-covariance matrix is defined as 
$$
S_{12}(t) := \nabla_{t_2} \nabla_{t_1} K_x(t).
$$ 
Its gradient with respect to $t_1$ is
$$
\begin{aligned}
\nabla_{t_1} K_x(t) & = \frac{ D_1\ebb ({\alpha e^{\alpha^{\top} h(t) } }) }{ \ebb ({ e^{\alpha^{\top} h(t) } }) } 
+ \frac{ \ebb ({ \varepsilon_1 e^{t_1^{\top} \varepsilon_1 } }) }{ \ebb ({ e^{t_1^{\top} \varepsilon_1 } }) },
\end{aligned}
$$
where the last term does not depend on $t_2$. Computing the gradient of this expression with respect to $t_2$ gives
$$
\begin{aligned}
S_{12}(t) = D_1 \ccal_{\alpha} ({ h(t) }) \Rra{ \diag [{ e^{t_2} }] D_2 }^{\top},
\end{aligned}
$$
where we substituted expression~\eqref{gen-covariance-alpha} for the generalized covariance of the independent sources.

\subsubsection{Approximation of the T-cumulants with the generalized covariance matrix}
\label{app:dir-derivative-approx}

Let $f_{mm'}(t) = [\ccal_x(t)]_{mm'}$ be a function $\R\to \R^M$ corresponding to the $(m,m')$-th element of the generalized covariance matrix. Then the following holds for its directional derivative at $t_0$ along the direction $t$:
$$
\inner{\nabla f_{mm'}(t_0),t} = \lim_{\delta\to0}\frac{ f_{mm'}(t_0+\delta t) - f_{mm'}(t_0)}{\delta},
$$
where $\inner{\cdot,\cdot}$ stands for the inner product. Therefore, when using the fact that $\nabla f(t_0) = \nabla \ccal_x(t)$ is the generalized cumulant of $x$ at $t_0$ and the definition of a projection of a tensor onto a vector~\eqref{projection}, one obtains for $t_0=\zero$ the approximation of the cumulant $\cum(x)$ with the generalized covariance matrix $\ccal_x(t)$.

Let us define $v_1 = W_1^{\top} u_1$ and $v_1 = W_2^{\top} u_2$ for some $u_1,\,u_2 \in\R^K$. Then, approximations for the T-cumulants~\eqref{Tdcca} of discrete CCA take the following form:
$W_1T_{121}(v_1)W_2$ is approximated by the generalized S-covariances~\eqref{GenSdcca} $S_{12}(t)$ via the following expression
$$
\begin{aligned}
W_1T_{121}(v_1)W_2 & \approx \frac{ W_1 S_{12} (\delta t_1) W_2^{\top} - W_1 S_{12}(\zero) W_2^{\top} }{\delta} \\
& - W_1 \diag(v_1) S_{12} W_2^{\top},
\end{aligned}
$$
where $t_1 = \begin{pmatrix} v_1 \\ 0 \end{pmatrix}$ and $W_1T_{122}(v_2)W_2$ is approximated by the generalized S-covariances $S_{12}(t)$ via
$$
\begin{aligned}
W_1T_{122}(v_2)W_2 & \approx \frac{ W_1 S_{12} (\delta t_2) W_2^{\top} - W_1 S_{12}(\zero) W_2^{\top} }{\delta} 
\\ & - W_1  S_{12} \diag(v_2) W_2^{\top},
\end{aligned}
$$
where $t_2 = \begin{pmatrix} 0 \\ v_2 \end{pmatrix}$ and $\delta$ are chosen to be small.

\subsection{Finite sample estimators}
\label{app:finite-sample-estimators}
\subsubsection{Finite sample estimators of the generalized expectation and covariance matrix}
\label{app:finite-sample-generalized-covariances}

Following \citet{Yer2000,SlaYer2012b}, we use the most direct way of defining the finite sample estimators of the generalized expectation~\eqref{gen-expectation}  and covariance matrix~\eqref{gen-covariance}.

Given a finite sample $X=\cbra{x_1,x_2, \dots,x_N}$, an estimator of the generalized expectation is
$$
\wh{\ecal}_x (t) = \frac{ \sumn x_n w_n }{\sumn w_n }
$$
where weights $w_n = e^{t^{\top} x_n}$
and an estimator of the generalized covariance is
$$
\wh{\ccal}_x (t) = \frac{ \sumn x_n x_n^{\top} w_n }{\sumn w_n } - \wh{\ecal}_x(t) \wh{\ecal}_x(t)^{\top}.
$$
Similarly, an estimator of the generalized S-covariance matrix is then
$$
\begin{aligned}
\wh{\ccal}_{x_1,x_2} (t) &= \frac{ \sumn x_{1n} x_{2n}^{\top} w_n }{\sumn w_n }
 - \frac{ \sumn x_{1n} w_n }{ \sumn w_n } \frac{ \sumn x_{2n}^{\top} w_n }{ \sumn w_n },
\end{aligned}
$$
where $x=[x_1;\;x_2]$ and $t=[t_1;\;t_2]$ for some $t_1\in\R^{M_1}$ and $t_2\in\R^{M_2}$.

Some properties of these estimators are analyzed by \citet{SlaYer2012b}.

\subsubsection{Finite sample estimators of the DCCA cumulants}
\label{app:finite-sample-cumulants}
In this section, we sketch the derivation of unbiased finite sample estimators for the CCA cumulants $S_{12}$, $T_{121}$, and $T_{122}$. Since the derivation is nearly identical to the derivation of the estimators for the DICA cumulants (see Appendix F.2 of \citet{PodEtAl2015}), all details are omitted.

Given a finite sample $X_1=\cbra{ x_{11}, x_{12},\dots,x_{1N} }$ and $X_2=\cbra{x_{21}, x_{22},\dots,x_{2N}}$, the finite sample estimator of the discrete CCA S-covariance~\eqref{Sdcca}, i.e., $S_{12} := \cum(x_1,x_2)$, takes the form 
\begin{equation}
\label{temp:sample-estimator-s12}
\wh{S}_{12}  = \eta_1 \sbra{ X_1X_2^{\top} - N \webb(x_1) \webb(x_2)^{\top} },
\end{equation}
where $\webb(x_1)=N^{-1}\sumn x_{1n}$, $\webb(x_2) = N^{-1}\sumn x_{2n}$, and $\eta_1 = 1/(N-1)$.

Substitution of the finite sample estimators of the 2nd and 3rd cumulants (see, e.g., Appendix C.4 of \citet{PodEtAl2015}) into the definition of the DCCA T-cumulants~\eqref{Tdcca} leads to the following expressions
\begin{align*}
\wh{W}_1&\wh{T}_{12j}(v_j)\wh{W}_2^{\top} = 
\eta_2 [(\wh{W}_1 X_1) \diag( X_j^{\top} v_j )]\otimes (\wh{W}_2 X_2) \\
&+ \eta_2 \innerp{v_j,\webb(x_j)}  2N [\wh{W}_1\webb(x_1)] \otimes [\wh{W}_2 \webb(x_2)] \\
&- \eta_2 \innerp{v_j,\webb(x_j)} (\wh{W}_1X_1) \otimes (\wh{W}_2 X_2)  \\
&-\eta_2  [ (\wh{W}_1 X_1) (X_j^{\top} v_j) ] \otimes [\wh{W}_2\webb(x_2)] \\
&- \eta_2 [\wh{W}_1\webb(x_1)] \otimes [ (\wh{W}_2 X_2)(X_j^{\top} v_j) ]
 \\
& - \eta_1 (\wh{W}_1^{(j)} X_1) \otimes (\wh{W}_2^{(j)} X_2) \\
& + \eta_1 N [\wh{W}_1^{(j)} \webb(x_1)] \otimes [\wh{W}_2^{(j)} \webb(x_2)],
\end{align*}
where $\eta_2 = N / ( (N-1)(N-2) )$ and $\wh{W}_1^{(1)} = \wh{W}_1\diag(v_1)$, $\wh{W}_2^{(1)}=\wh{W}_2$, $\wh{W}_1^{(2)}=\wh{W}_1$, and $\wh{W}_2^{(2)}=\wh{W}_2\diag(v_2)$.

In the expressions above, $\wh{W}_1$ and $\wh{W}_2$ denote whitening matrices of $\wh{S}_{12}$, i.e. such that $\wh{W}_1 \wh{S}_{12} \wh{W}_2^{\top} = I$.

\subsection{Implementation details}
\label{app:implementation-details}

\subsubsection{Construction of S- and T-cumulants}
\label{app:S-and-T-cumulants}
By analogy with \citet{PodEtAl2015}, the target matrices for joint diagonalization can be constructed from S- and T-cumulants.

When dealing with the S- and T-cumulants, the target matrices are obtained via tensor projections. We define a projection $\tcal(v)\in\R^{M_1\times M_2}$ of a third-order tensor $\tcal\in\R^{M_1\times M_2\times M_3}$ onto a vector $v\in\R^{M_3}$ as
\begin{equation}\label{projection}
[\tcal(v)]_{m_1m_2} := \sum_{m_3=1}^{M_3} [\tcal]_{m_1m_2m_3} v_{m_3}.
\end{equation}
Note that the projection $\tcal(v)$ is a matrix.
Therefore, given $2P$ vectors $\cbra{v_{11},v_{21}, v_{12}, v_{22}, \dots,v_{1P},v_{2P}}$, one can construct $2P + 1$ matrices 
\begin{equation}\label{jdMatricesS&T}
\{S_{12}, \;\; T_{121}(v_{1p}), \;\; T_{122}(v_{2p}), \;\; \mbox{ for } p=1,\dots, P\},
\end{equation}
which have the diagonal form~\eqref{diagSdcca} and~\eqref{diagTdcca}. 
Importantly, the tensors are never constructed (see \citet{AnaEtAl2012,AnaEtAl2014,PodEtAl2015} and Appendix~\ref{app:finite-sample-cumulants}).

\subsubsection{Computation of whitening matrices}
\label{app:compute-whitening}
One can compute such whitening matrices~\eqref{whitening} via the singular value decomposition (SVD) of $S_{12}$. Let $S_{12} = U \Sigma V^{\top}$ be the SVD of $S_{12}$, then one can define $W_1 = U_{1:K} \Lambda$ and $W_2 = V_{1:K} \Lambda$, where $U_{1:K}$ and $V_{1:K}$ are the first $K$ left- and right-singular vectors and $\Lambda=\diag( \sigma_1^{-1/2},\dots,\sigma_K^{-1/2})$ and $\sigma_1,\dots,\sigma_K$ are the $K$ largest singular values. 

Although SVD is computed only once, the size of the matrix $S_{12}$ can be significant even for storage. To avoid construction of this large matrix and speed up SVD, one can use randomized SVD techniques \citep{HalEtAl2011}. Indeed, since the sample estimator $\wh{S}_{12}$ has the form~\eqref{temp:sample-estimator-s12}, one can reduce this matrix by sampling two Gaussian random matrices $\Omega_1 \in\R^{\tilde{K} \times M_1}$ and $\Omega_2 \in\R^{\tilde{K} \times M_2}$, where $\tilde{K}$ is slightly larger than $K$. Now, if $U$ and $V$ are the $K$ largest singular vectors of the reduced matrix $\Omega_1 \wh{S}_{12} \Omega_2$, then $\Omega_1^{\dagger} U$ and $\Omega_2^{\dagger} V$ are approximately (and up to permutation and scaling of the columns) the $K$ largest singular vectors of $\wh{S}_{12}$.

\subsubsection{Applying whitening transform to DCCA T-cumulants}
\label{app:whiten-t-cumulants}
Transformation of the T-cumulants~\eqref{jdMatricesS&T} with whitening matrices $W_1$ and $W_2$ gives new tensors $\wh{T}_{12j} \in\R^{K\times K\times K}$:
\begin{equation}\label{rrrr}
\wh{T}_{12j} := T_{12j} \times_1 W_1^{\top} \times_2 W_2^{\top} \times_3 W_j^{\top},
\end{equation}
where $j=1,2$. Combining this transformation with the projection~\eqref{projection}, one obtains $2P+1$ matrices
\begin{equation}\label{jdmatricesS&Twhitened}
W_1S_{12}W_2^{\top},\; W_1 T_{12j}(W^{\top}_j u_{jp}) W_2^{\top},
\end{equation}
where $p=1,\dots,P$ and $j=1,2$ and we used $v_{jp} = W_j^{\top} u_{jp}$ to take into account whitening along the third direction. By choosing $u_{jp}\in\R^K$ to be the canonical vectors of the $R^K$, the number of tensor projections is reduced from $M=M_1+M_2$ to $2K$.

\subsubsection{Choice of projection vectors or processing points}
\label{app:processing-projection-points}

For the T-cumulants~\eqref{jdMatricesS&T}, we choose the $K$ projection vectors as $v_{1p}=W_1^{\top}e_p$ and $v_{2p}=W_2^{\top}e_p$, where $e_p$ is one of the columns of the $K$-identity matrix (i.e., a canonical vector). For the generalized S-covariances~\eqref{jdmatricesGenS}, we choose the processing points as $t_{1p} = \delta_1 v_{1p}$ and $t_{2p} = \delta_2 v_{2p}$, where $\delta_j$, for $j=1,2$ are set to a small value such as $0.1$ divided by $\sum_m\ebb(|x_{jm}|)/M_j$, for $j=1,2$.

When projecting a tensor $T_{12j}$ onto a vector, part of the information contained in this tensor gets lost. To preserve all information, one could project a tensor $T_{12j}$ onto the canonical basis of $\R^{M_j}$ to obtain $M_j$ matrices. However, this would be an expensive operation in terms of both memory and computational time. In practice, we use the fact, that the tensor $T_{12j}$, for $J=1,2$, is transformed with whitening matrices~\eqref{rrrr}. Hence, the projection vector has to include multiplication by the whitening matrices. Since they reduce the dimension to $K$, choosing the canonical basis in $\R^{K}$ becomes sufficient. Hence, the choice $v_{1p} = W_1^{\top} e_p$ and $v_{2p} = W_2^{\top} e_p$, where $e_p$ is one of the columns of the $K$-identity matrix.

Importantly, in practice, the tensors are never constructed (see Appendix~\ref{app:finite-sample-cumulants}). 

The choice of the processing points of the generalized covariance matrices has to be done carefully. Indeed, if the values of $t_1$ or $t_2$ are too large, the exponents blow up. Hence, it is reasonable to maintain the values of the processing points very small. Therefore, for $j=1,2$, we set $t_{jp} = \delta_j v_{jp}$ where $\delta_j$ is proportional to a parameter $\delta$ which is set to a small value ($\delta = 0.1$ by default), and the scale is determined by the inverse of the empirical average of the component of $x_j$, i.e.: 
\begin{equation} \label{eq:delta_j}
\delta_j := \delta \frac{ N M_j }{\sum_{n=1}^N \sum_{m=1}^{M_j} [ |X_j| ]_{mn}},
\end{equation}
for $j=1,2$. See Appendix~\ref{app:gdcca-vs-delta} for an experimental comparison of different values of $\delta$ (the default value used in other experiments is $\delta=0.1$).

\subsubsection{Finalizing estimation of $D_1$ and $D_2$}
\label{app:est-D}

The non-orthogonal joint diagonalization algorithm outputs an invertible matrix $Q$. If the estimated factor loading matrices are not supposed to be non-negative (continuous case of NCCA~\eqref{ncca}), then
\begin{equation}\label{temp:finalizing-estimation}
\begin{aligned}
D_1 &= W_1^{\dagger} Q, \\
D_2 &= W_2^{\dagger} Q^{-1},
\end{aligned}
\end{equation}
where $\dagger$ stands for the pseudo-inverse.
For the spectral algorithm, where $Q$ are eigenvectors of a non-symmetric matrix and are not guaranteed to be real, only real parts are kept after evaluating matrices $D_1$ and $D_2$ in accordance with~\eqref{temp:finalizing-estimation}.

If the matrices $D_1$ and/or $D_2$ have to be non-negative (the discrete case of DCCA~\eqref{dcca} and MCCA~\eqref{mcca}), they have to be further mapped. For that, we select the sign of each column such that the vector (column) has less negative than positive components, which is measured by the sum of squares of the components of each sign, (this is necessary since the scaling unidentifiability includes the scaling by $-1$) and then truncate all negative values at 0. 

In practice, due to the scaling unidentifiability, each column of the obtained matrices $D_1$ and $D_2$ can be further normalized to have the unit $\ell_1$-norm. This is applicable in all cases (D/M/NCCA).

\subsection{Jacobi-like joint diagonalization of non-symmetric matrices}
\label{app:nojd}

Given $N$ non-defective (a.k.a.~diagonalizable) not necessary \emph{normal}\footnote{A real matrix $A$ is normal if $A^{\top}A=AA^{\top}$.} matrices 
$$
\acal = \cbra{A_1,\, A_2,\, \dots,\, A_N},
$$ 
where each matrix $A_n\in\R^{M\times M}$, find such matrix $Q\in\R^{M\times M}$ that matrices 
$$
Q^{-1}\acal Q = \cbra{Q^{-1}A_1Q, \, Q^{-1}A_2Q,\, \dots,\,Q^{-1}A_NQ}
$$ 
are (jointly) as diagonal as possible. We refer to this problem as a non-orthogonal joint diagonalization (NOJD) problem.\footnote{An orthogonal joint diagonalization problem corresponds to the case where the matrices $A_1,\,A_2,\,\dots,\,A_N$ are normal and, hence, diagonalizable by an orthogonal matrix $Q$.}

\begin{algorithm}[!ht]
   \caption{Non-orthogonal joint diagonalization (NOJD)}
   \label{alg:nojd}
\begin{algorithmic}[1]
	\STATE Initialize: $\acal^{(0)}\leftarrow \acal$ and $Q^{(0)} \leftarrow I_M$ and iterations $\ell=0$
	\FOR {sweeps $k=1,2,\dots$}
		\FOR {$p=1,\,\dots,\,M-1$}
			\FOR {$q=p+1,\,\dots,\,M$}
			    \STATE Increase $\ell= \ell + 1$
				\STATE Find the (approx.) shear parameter $y^*$ defined in~\eqref{shear-opt} 
				\STATE Find the Jacobi angle $\theta^*$ defined in~\eqref{unitary-opt} 
				\STATE Update $Q^{(\ell)} \leftarrow Q^{(\ell-1)}S^{(\ell)}_{*}U_*^{(\ell)}$
				\STATE Update $\acal^{(\ell)} \leftarrow U^{(\ell)\top}_{*}S^{(\ell)-1}_{*} \acal^{(\ell-1)} S^{(\ell)}_{*}U^{(\ell)}_{*}$ \label{line:Aupdate}
			\ENDFOR
		\ENDFOR
	\ENDFOR
	\STATE Output: $Q^{(\ell)}$
\end{algorithmic}
\end{algorithm}

\textbf{Algorithm}. Non-orthogonal {J}acobi-like joint diagonalization algorithms have the high level structure which is outlined in Alg.~\ref{alg:nojd}.

The algorithm iteratively constructs the sequence of matrices $\acal^{(\ell)}=\cbra{A_1^{(\ell)},\,A_2^{(\ell)},\,\dots,\,A_N^{(\ell)}}$, which is initialized with $\acal^{(0)}=\acal$. Each such iteration $\ell$ corresponds to a single update (Line~\eqref{line:Aupdate} of Alg.~\eqref{alg:nojd}) of the matrices with the optimal shear $S^{(\ell)}_{*}$ and unitary $U^{(\ell)}_{*}$ transforms:
$$
A_n^{(\ell)} = U^{(\ell)\top}_{*} S^{(\ell)-1}_{*} A_n^{(\ell-1)} S^{(\ell)}_{*}U^{(\ell)}_{*},
$$
where $S^{(\ell)}_{*}=S^{(\ell)}(y^*)$ and $U^{(\ell)}_{*}=U^{(\ell)}(\theta^*)$ for the chosen in accordance with some rules (see below) optimal shear parameter $y^*$ and optimal Jacobi (=Givens) angle $\theta^*$.

For the theoretical analysis purposes, the two transforms are considered separately:
\begin{equation}\label{Aprimes}
\begin{aligned}
A'^{(\ell)}_n &= S^{(\ell)-1}(y) A^{(\ell-1)}_n S^{(\ell)}(y), \\
A^{(\ell)}_n & = A''^{(\ell)}_n = U^{(\ell)\top}(\theta) A'^{(\ell)}_n U^{(\ell)}(\theta).
\end{aligned}
\end{equation}

Each such iteration $\ell$ is a combination of the iteration $k$ and the pivots $p$ and $q$ (see Alg.~\ref{alg:nojd}). The iteration $k$ is referred to as a \emph{sweep}. Within each sweep $k$, $M(M-1)/2$ pivots $p<q$ are chosen in accordance with the lexicographical rule. The rule for the choice of pivots can affect convergence as was analyzed for the single matrix case \citep[see, e.g.,][]{Ruh1968,Ebe1962}, where more sophisticated rules were proposed for the algorithm to have a quadratic convergence phase. However, up to our best knowledge, no such analysis was done for the several matrices case. We assume the simple lexicographical rule all over the paper.

The \emph{shear transform} is defined by the hyperbolic rotation matrix $S^{(\ell)} = S^{(\ell)}(y)$ which is equal to the identity matrix except for the following entries
\begin{equation}\label{shear}
\begin{pmatrix}
S^{(\ell)}_{pp} & S^{(\ell)}_{pq} \\
S^{(\ell)}_{qp} & S^{(\ell)}_{qp} 
\end{pmatrix}
=
\begin{pmatrix}
\cosh y & \sinh y \\
\sinh y & \cosh y
\end{pmatrix},
\end{equation}
where the \emph{shear parameter} $y\in\R$.
The \emph{unitary transform} is defined by the Jacobi (=Givens) rotation matrix $U^{(\ell)}= U^{(\ell)}(\theta)$ which is equal to the identity matrix except for the following entries
\begin{equation}\label{unitary}
\begin{pmatrix}
U^{(\ell)}_{pp} & U^{(\ell)}_{p	q} \\
U^{(\ell)}_{qp} & U^{(\ell)}_{qp} 
\end{pmatrix}
=
\begin{pmatrix}
\cos \theta & \sin \theta \\
-\sin \theta  & \cos \theta
\end{pmatrix},
\end{equation}
where the \emph{Jacobi (=Givens)} angle $\theta\in\sbra{-\frac{\pi}{4}, \frac{\pi}{4}}$.

The following two objective functions are of the central importance for this type of algorithms: (a) the sum of squares of all the off-diagonal elements of the matrices\footnote{In the JUST algorithm \citep{IfeEtAl2009}, this objective function is also considered for the (shear transformed) matrix $\acal'^{(\ell)}$.} $\acal''^{(\ell)}$ which are the transformed with the unitary transform $U^{(\ell)}$  matrices $\acal'^{(\ell)}$:
\begin{equation}
\label{obj-off}
\off\rbra{\acal''^{(\ell)}} = \sum_{n=1}^N \off \rbra{ U^{(\ell)\top}  A'^{(\ell)}_n U^{(\ell)} }
\end{equation}
and (b) the sum of the squared Frobenius norms of the matrices $\acal'^{(\ell)}$ which are the transformed with the share transform $S^{(\ell)}$ matrices $\acal^{(\ell-1)}$:
\begin{equation}\label{obj-fro}
\norm{\acal'^{(\ell)}}_F^2 = \sum_{n=1}^N \norm{ S^{(\ell)-1}A_n^{(\ell-1)}S^{(\ell)}}_F^2.
\end{equation}
We refer to~\eqref{obj-off} as the \emph{diagonality measure} and to~\eqref{obj-fro} as the \emph{normality measure}.

All the considered algorithms find the optimal Jacobi angle $\theta^*$ as the minimizer of the diagonality measure of the (unitary transformed) matrices $\acal''^{(\ell)}$~\eqref{Aprimes}:
\begin{equation}\label{unitary-opt}
\theta^* = \argmin_{\theta\in[-\frac{\pi}{4}, \frac{ \pi}{4}]} \off \rbra{ \acal''^{(\ell)} },
\end{equation}
which admits a unique closed form solution \citep{CarSou1996}.
The optimal shear parameter $y^*$ is found%
\footnote{The JUST algorithm is an exception here, since it minimizes the diagonality measure $\off[{ \acal'^{(\ell)}}]$ of the (shear transformed) matrices $\acal'^{(\ell)}$ with respect to $y$.
}
as a minimizer of the normality measure of the (shear transformed) matrices $\acal'^{(\ell)}$~\eqref{Aprimes}:
\begin{equation}\label{shear-opt}
y^* = \argmin_{y\in\R} \norm{\acal'^{(\ell)}}_F^2.
\end{equation}
All the considered algorithms \citep{FuGao2006,IfeEtAl2009,LucAlb2010} solve this step only approximately. In particular, the sh-rt algorithm \citep{FuGao2006} approximates the equation for finding the nulls of the gradient of the objective; the JUST algorithm \citep{IfeEtAl2009} replaces the normality measure with the diagonality measure and provides a closed form solution for the resulting problem; and the JDTM algorithm \citep{LucAlb2010} replaces the normality measure with the sum of only two squared elements $A'_{n,pq}$ and $A'_{n,qp}$ and provides a closed form solution for the resulting problem.

The three NOJD algorithms can have slightly different convergence properties, however, for the purposes of this paper their performance can hardly be distinguished. That is, the difference in the performance of the algorithms in terms of the $\ell_1$-error of the factor loading matrices is hardly noticeable. For the experiments, we use the JDTM algorithm, the other two algorithms could be equally used. To the best of our knowledge, no theoretical analysis of the NOJD algorithms is available, except for the single matrix case when they boil down to the (non-symmetric) eigenproblem \citep{Ebe1962,Ruh1968}.

The following intuitively explains why the normality measure, i.e. the sum of the squared Frobenius norms, has to be minimized at the shear transform. As \citep{Ruh1968} mention, for every matrix $A$ and non-singular $Q$:
$$
\inf_Q \norm{Q^{-1}AQ}_F^2 = \norm{\Lambda}_F^2,
$$
where $\Lambda$ is the diagonal matrix containing the eigenvalues of $A$. Therefore, a diagonalized version of the matrix $A$ must have the smallest Frobenius norm. Since the unitary transform does not change the Frobenius norm, it can only be minimized with the shear transform. Further, if a matrix is normal, i.e. $A^{\top}A=AA^{\top}$ with a symmetric matrix as a particular case, the upper triangular matrix in its Schur decomposition is zero \citepsup[Chapter 7]{GolVan1996} and then the Schur vectors correspond to the (orthogonal in this case) eigenvectors of this matrix. Therefore, a normal non-defective matrix can be diagonalized by an orthogonal matrix, which preserves the Frobenius norm. Hence, the shear transform by minimizing the normality measure decreases the deviation from normality and then the unitary transform by minimizing the diagonality measure decreases the deviation from diagonality.

\subsection{Supplementary experiments}
\label{app:supplementary-experiments}

\begin{figure*}[!t]
\centering 
\begin{tabular}{ccc}
\includegraphics[width=.3\textwidth,clip=true, trim=5 5 5 5]{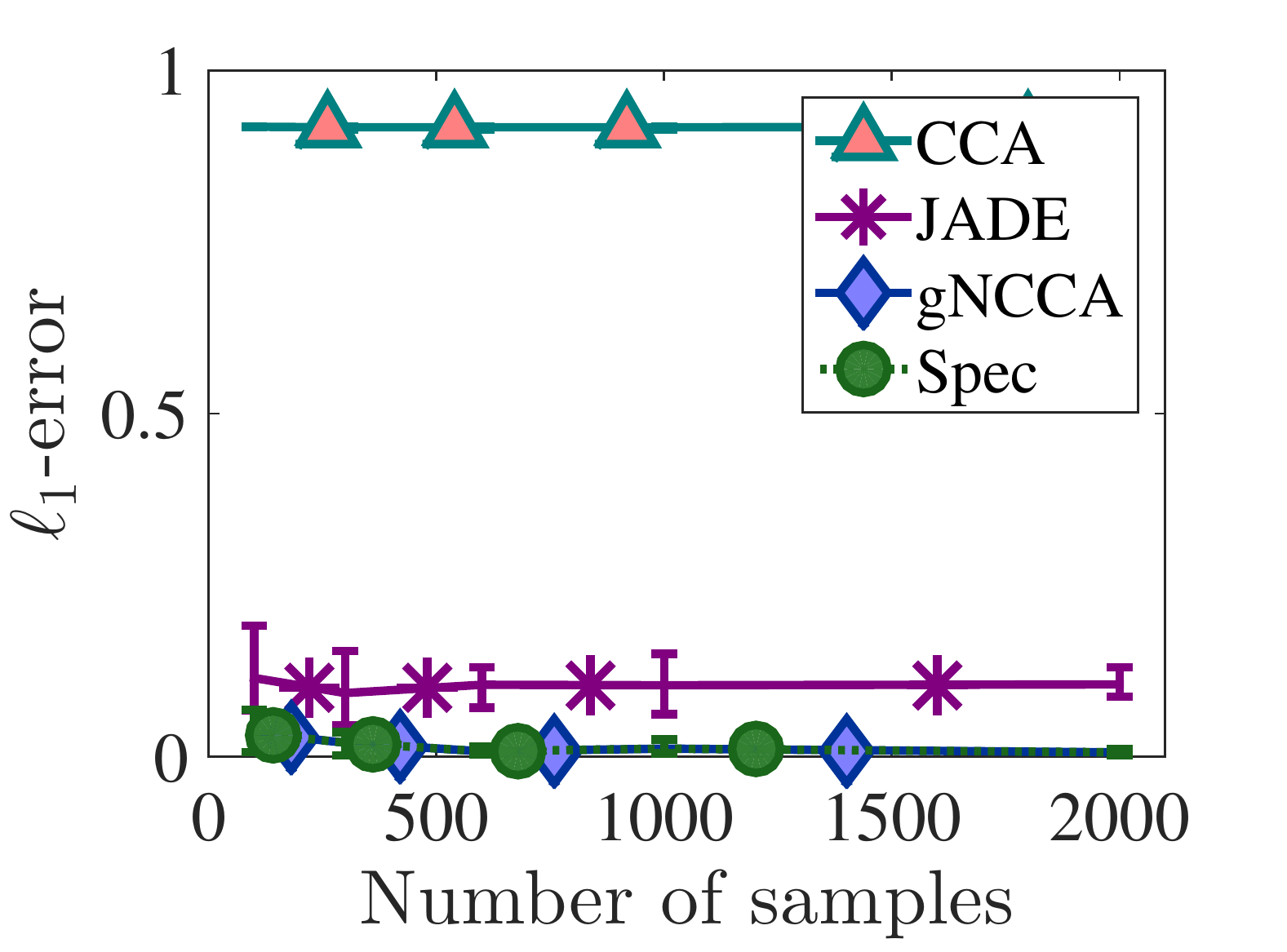} &
\includegraphics[width=.3\textwidth,clip=true, trim=5 5 5 5]{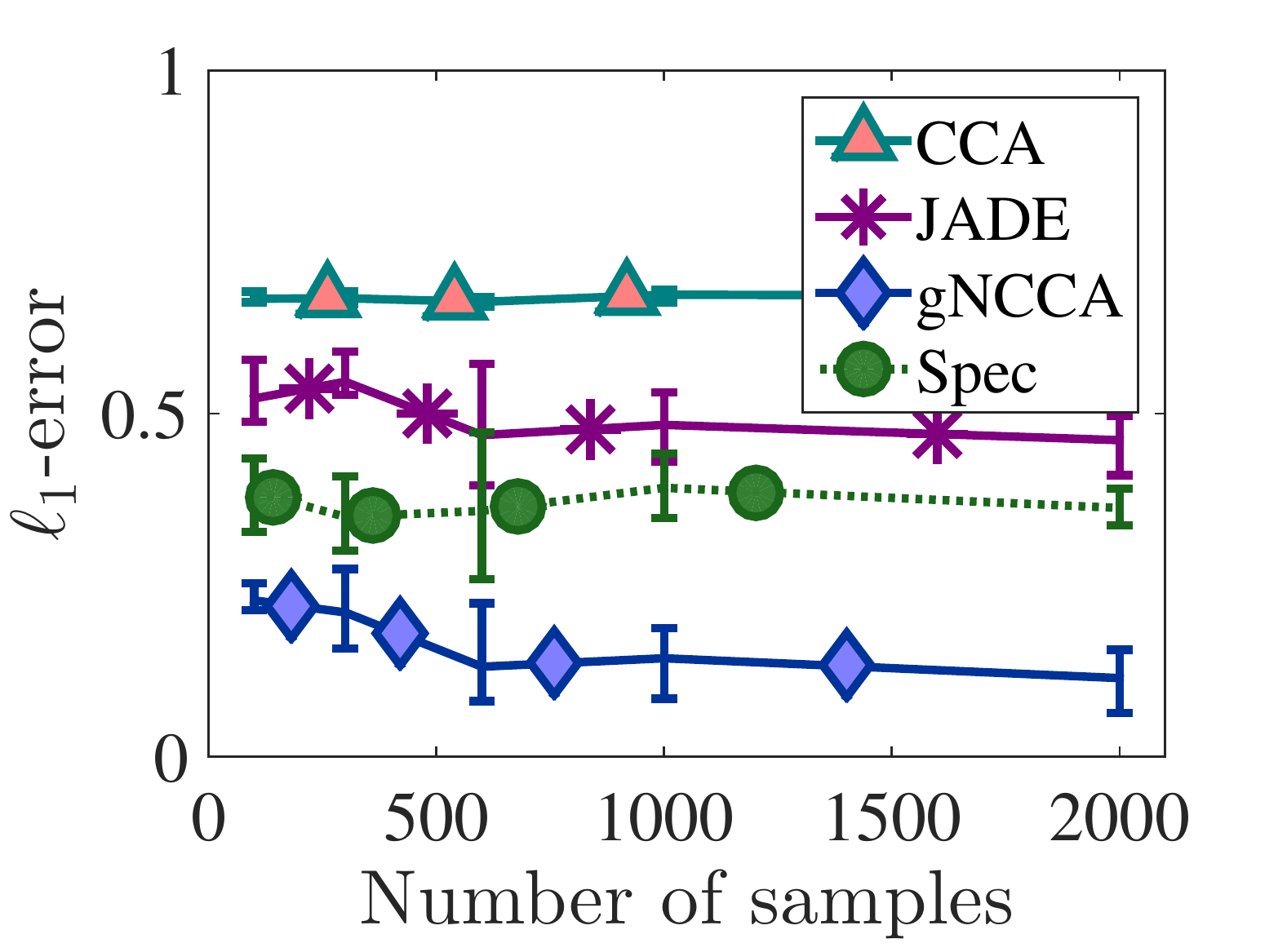} &
\includegraphics[width=.3\textwidth,clip=true, trim=5 5 5 5]{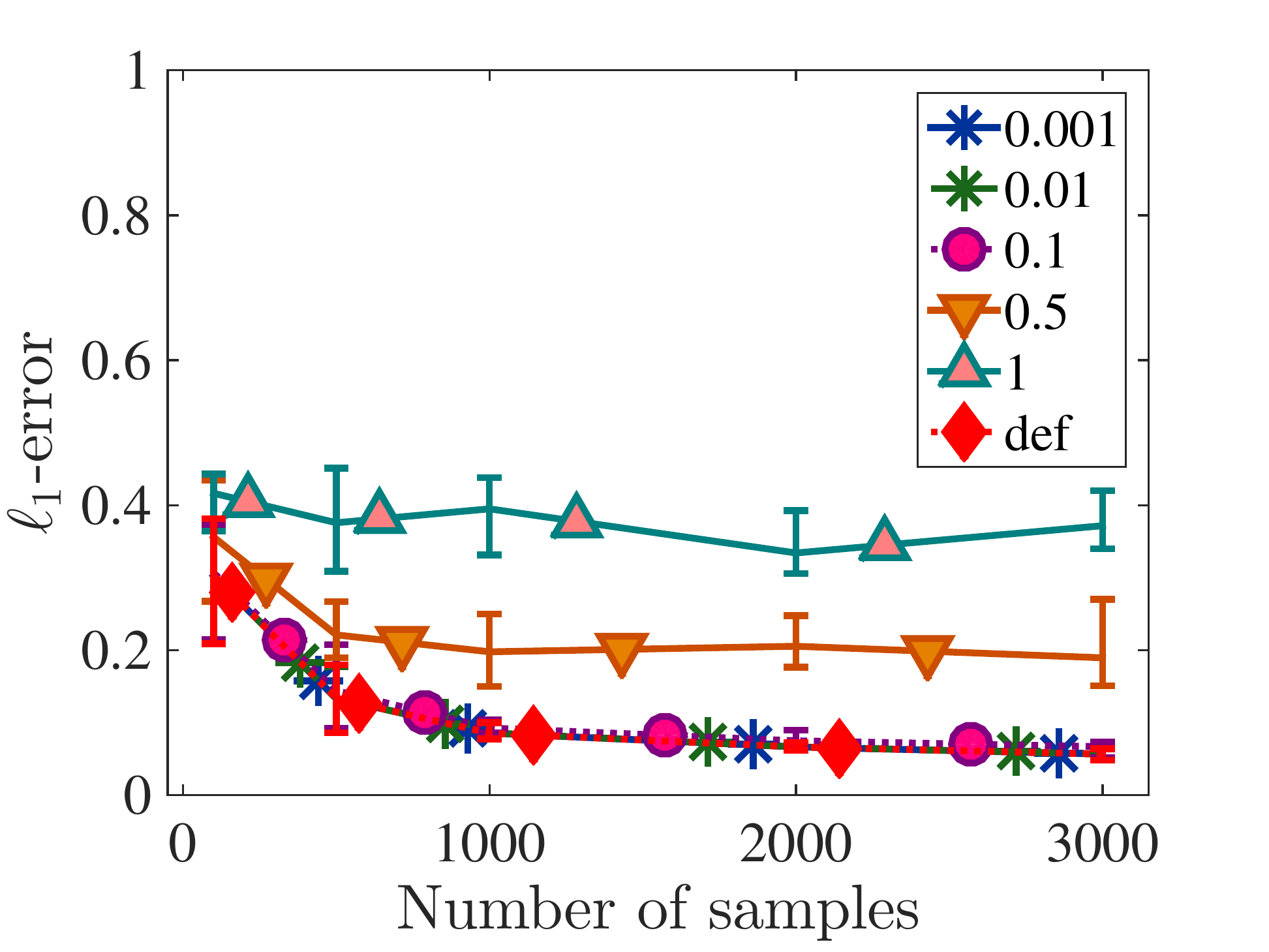}
\end{tabular}
\vspace{-1em}
\caption{\textbf{(left and middle)} The continuous synthetic data experiment from Appendix~\ref{sec:exp-continuous}  with $M_1=M_2=20$, $c=c_1=c_2=0.1$ and $L_n=L_s=1000$. The number of factors: \textbf{(left)} $K_1=K_2=K=1$  and \textbf{(middle)} $K_1=K_2=K=10$. \textbf{(right)}: An experimental analysis of the performance of DCCAg with generalized covariance matrices using different parameters $\delta_j$ for the processing points. The numbers in the legend correspond to the values of $\delta$ defining $\delta_j$ via~\eqref{eq:delta_j} in Appendix~\ref{app:processing-projection-points}. The default value (def) is $\delta = 0.1$.
The data is the discrete synthetic data as described in Section~\ref{sec:experiments} with the parameters set as in Fig.~\ref{fig:dica} (right).
\vspace{-1.5em}
}
\label{fig:cca-exp}
\end{figure*}
\setlength{\textfloatsep}{11pt}

The code for reproducing the experiments described in Section~\ref{sec:experiments} as well as in this appendix is available at \href{https://github.com/anastasia-podosinnikova/cca}{https://github.com/anastasia-podosinnikova/cca}.

\subsubsection{Continuous synthetic data}
\label{sec:exp-continuous}
This experiment is essentially a continuous analogue to the synthetic experiment with the discrete data from Section~\ref{sec:experiments}.

\ppp{Synthetic data.} We sample synthetic data from the linear non-Gaussian CCA (NCCA) model~\eqref{linear-cca} with each view $x_j = D_j \alpha + F_j \beta_j$. The (non-Gaussian) sources are $\alpha \sim z_{\alpha} \gam(c,b)$, where $z_{\alpha}$ is a Rademacher random variable (i.e., takes the values $-1$ or $1$ with the equal probabilities). The noise sources are $\beta_j \sim z_{\beta_j} \gam(c_j,b_j)$, for $j=1,2$,  where again $z_{\beta_j}$ is a Rademacher random variable. Parameters of the gamma distribution are initialized by analogy with the discrete case (see Section~\ref{sec:experiments}). The elements of the matrices $D_j$ and $F_j$, for $j=1,2$, are sampled i.i.d. for the uniform distribution in $[-1, \, 1]$. Each column of $D_j$ and $F_j$, for $j=1,2$, is normalized to have the unit $\ell_1$-norm.

\ppp{Algorithms.} We compare gNCCA (the implementation of NCCA with the generalized S-covariance matrices with the default values of the parameters $\delta_1$ and $\delta_2$ as described in Appendix~\ref{app:processing-projection-points}) the spectral algorithm for NCCA (also with the generalized S-covariance matrices) to the JADE algorithm 
\citep[the code is available at 
\href{http://perso.telecom-paristech.fr/~cardoso/Algo/Jade/jadeR.m}{http://perso.telecom-paristech.fr/~cardoso/Algo/Jade/jadeR.m;}][]{CarSou1993}
for independent component analysis (ICA) and to classical CCA.

\ppp{Synthetic experiment.} In Fig.~\ref{fig:cca-exp} (left and middle), the results of the experiment for the different number of topics are presented. The error of the classical CCA is high due to the mentioned unidentifiability issues.

\subsubsection{Sensitivity of the generalized covariance matrices to the choice of the processing points}
\label{app:gdcca-vs-delta}
In this section, we experimentally analyze the performance of the DCCAg algorithm based on the generalized S-covariance matrices vs. the parameters $\delta_1$ and $\delta_2$. We use the experimental setup of the synthetic discrete data from Section~\ref{sec:experiments} with $K_1=K_2=K=10$. The results are presented in Fig.~\ref{fig:cca-exp} (right).

\setlength{\textfloatsep}{9pt}
\begin{table*} 
\centering 
\begin{tabular}{cc|cc|cc|cc} 
              farmers &         agriculteurs &              division &                   no &                  nato &                 otan &                   tax &              imp\^ots  \\ 
           agriculture &            programme &             negatived &                 vote &                kosovo &               kosovo &                budget &               budget  \\ 
               program &             agricole &                paired &             rejet\'ee &                forces &           militaires &               billion &              enfants  \\ 
                  farm &                 pays &               declare &                 voix &              military &               guerre &              families &            \'economie  \\ 
               country &            important &                  yeas &                 mise &                   war &        international &                income &              ann\'ees  \\ 
               support &            probl\`eme &               divided &             pairs &                troops &                 pays &               country &              dollars  \\ 
              industry &                 aide &                  nays &                porte &               country &           r\'efugi\'es &                  debt &                 pays  \\ 
                 trade &          agriculture &                  vote &               contre &                 world &            situation &              students &             finances  \\ 
             provinces &              ann\'ees &                 order &         d\'eclaration &              national &                 paix &              children &             familles  \\ 
                  work &              secteur &                deputy &           suppl\'eant &                 peace &          yougoslavie &                 money &               fiscal  \\ 
               problem &            provinces &             thibeault &                 vice &         international &            milosevic &               finance &            milliards  \\ 
                 issue &                 gens &            mcclelland &           lethbridge &              conflict &               forces &             education &            lib\'eraux  \\ 
                    us &            \'economie &                    ms &              poisson &             milosevic &               serbes &               liberal &               jeunes  \\ 
                   tax &            industrie &               oversee &                  mme &                debate &         intervention &                  fund &                 gens  \\ 
                 world &              dollars &                  rise &              plantes &               support &              troupes &                  care &            important  \\ 
                  help &               mesure &                  past &               harvey &                action &          humanitaire &               poverty &               revenu  \\ 
               federal &                 faut &                  army &             perdront &              refugees &              nations &                  jobs &               mesure  \\ 
             producers &            situation &              peterson &             sciences &                ground &              conflit &              benefits &               argent  \\ 
              national &          r\'eformiste &                  heed &             libert\'e &                happen &             ethnique &                 child &               sant\'e  \\ 
              business &               accord &                 moral &              pri\`ere &                 issue &                monde &                   pay &                payer  \\ 
 \end{tabular}
 \vspace{-.5em}
\caption{ The real data (translation) experiment. Topics 1 to 4. \vspace{-1.5em}} 
\label{tab:1}
\end{table*}

\subsubsection{Real data experiment -- translation topics}
\label{app:translation}  

For the real data experiment, we estimate the factor loading matrices (topics, in the following) $D_1$ and $D_2$ of aligned proceedings of the 36-th Canadian Parliament in English and French languages (can be found at \href{http://www.isi.edu/natural-language/download/hansard/}{http://www.isi.edu/natural-language/download/hansard/}). 

Although going into details of natural language processing (NLP) related problems is not the goal of this paper, we do minor pre-processing (see Appendix~\ref{app:translation-preprocess}) of this text data to improve the presentation of the estimated bilingual topics $D_1$ and $D_2$.

The 20 topics obtained with DCCA are presented in Tables~\ref{tab:1}--\ref{tab:5}. For each topic, we display the 20 most frequent words (ordered from top to bottom in the decreasing order). Most of the topic have quite clear interpretation. Moreover, we can often observe the pairs of words which are each others translations in the topics. Take, e.g., 
\begin{mi}
\item 
the topic 10: the phrase ``pension plan'' can be translated as ``r\'egime de retraite'', the word ``benefits'' as ``prestations'', and abbreviations ``CPP'' and ``RPC'' stand for ``Canada Pension Plan'' and ``R\'egime de pensions du Canada'', respectively;
\item 
the topic 3: ``OTAN'' is the French abbreviation for ``NATO'', the word ``war'' is translated as ``guerre'', and the word ``peace'' as ``paix'';
\item
the topic 9: ``Nisga'' is the name of an Indigenous (or ``aboriginal'') people in British Columbia, the word ``aboriginal'' translates to French as ``autochtontes'', and, e.g., the word ``right'' can be translated as ``droit''.
\end{mi}
Note also that, e.g., in topic 10, although the French words ``ans'' and ``ann\'ees'' are present in the French topic, their English translation ``year'' is not, since it was removed as one of the 15 most frequent words in English (see Appendix~\ref{app:translation-preprocess}).

\subsubsection{Data preprocessing} 
\label{app:translation-preprocess}
For the experiment, we use House Debate Training Set of the Hansard collection, which can be found at \href{http://www.isi.edu/natural-language/download/hansard/}{http://www.isi.edu/natural-language/download/hansard/}.

To pre-process this text data, we perform case conversion, stemming, and removal of some stop words. For stemming, the SnowballStemmer of the NLTK toolbox by \citetsup{NLTK} was used for both English and French languages. Although this stemmer has particular problems (such as mapping several different forms of a word to a single stem in one language but not in the other), they are left beyond our consideration. Moreover, in addition to the standard stop words of the NLTK toolbox, we also removed the following words that we consider to be stop words for our task\footnote{This list of words was obtained by looking at words that appear in the top-20 words of a large number of topics in a first experiment. Removing these words did not change much the content of the topics, but made them much more interpretable.} (and their possible forms):
\begin{mi}
\item from English: ask, become, believe, can, could, come, cost, cut, do, done, follow, get, give, go, know, let, like, listen, live, look, lost, make, may, met, move, must, need, put, say, see, show, take, think, talk, use, want, will, also, another, back, day, certain, certainly, even, final, finally, first, future, general, good, high, just, last, long, major, many, new, next, now, one, point, since, thing, time, today, way, well, without;
\item from French (translations in brackets): demander (ask), doit (must), devenir (become), dit (speak, talk), devoir (have to), donner (give), ila (he has), met (put), parler (speak, talk), penser (think), pourrait (could), pouvoir (can), prendre (take), savoir (know), aller (go), voir (see), vouloir (want), actuellement, apr\`es (after), aujourd'hui (today), autres (other), bien (good), beaucoup (a lot), besoin (need), cas (case), cause, {c}ela (it), certain, chose (thing), d\'ej\`a (already), dernier (last), \'egal (equal), entre (between), fa{\c{c}}on (way), grand (big), jour (day), lorsque (when), neuf (new), pass\'e (past), plus, point, pr\'esent, pr\^ets (ready), prochain (next), quelque (some), suivant (next), unique.
\end{mi}

After stemming and removing stop words, several files had different number of documents in each language and had to be removed too. The numbers of these files are: 16, 36, 49 55, 88, 103, 110, 114, 123, 155, 159, 204, 229, 240, 2-17, 2-35. 

We also removed the 15 most frequent words from each language. These include:
\begin{mi}
\item in English: Mr, govern, member, speaker, minist(er), Hon, Canadian, Canada, bill, hous(e), peopl(e), year, act, motion, question;
\item in French: gouvern(er), pr\'esident, loi, d\'eput(\'e), ministr(e), canadien, Canada, projet, Monsieur, question, part(y), chambr(e), premi(er), motion, Hon.
\end{mi}
Removing these words is not necessary, but improves the presentation of the learned topics significantly. Indeed, the most frequent words tend to appear in nearly every topic (often in pairs in both languages as translations of each other, e.g., ``member'' and ``d\'eput\'e'' or ``Canada'' in both languages, which confirms one more time the correctness of our algorithm).

Finally, we select $M_1=M_2=5,000$ words for each language to form matrices $X_1$ and $X_2$ each containing $N=11,969$ documents in columns. As stemming removes the words endings, we map the stemmed words to the respective most frequent original words when showing off the topics in Tables~\ref{tab:1}-\ref{tab:5}.

\subsubsection{Running time} 
\label{app:running-time}
For the real experiment, the runtime of DCCA algorithm is 24 seconds including 22 seconds for SVD at the whitening step. In general, the computational complexity of the D/N/MCCA algorithms is bounded by the time of SVD plus $O(RNK) + O(NK^2)$, where $R$ is the largest number of non-zero components in the stacked vector $x=[x_1;\;x_2]$, plus the time of NOJD for $P$ target matrices of size $K$-by-$K$. In practice, DCCAg is faster than DCCA.

\bibliographysup{litsup}
\bibliographystylesup{icml2016}

\newpage
\cleardoublepage
\setlength{\textfloatsep}{9pt}
\begin{table*}[h!] 
\centering
\begin{tabular}{cc|cc|cc|cc} 
                 work &              travail &               justice &               jeunes &              business &          entreprises &                 board &           commission  \\ 
               workers &        n\'egociations &                 young &              justice &                 small &              petites &                 wheat &                 bl\'e  \\ 
                strike &         travailleurs &                 crime &             victimes &                 loans &            programme &               farmers &         agriculteurs  \\ 
           legislation &               gr\`eve &             offenders &             syst\'eme &               program &              banques &                 grain &       administration  \\ 
                 union &               emploi &               victims &                crime &                  bank &             finances &             producers &          producteurs  \\ 
             agreement &                droit &                system &               mesure &                 money &            important &             amendment &                grain  \\ 
                labour &             syndicat &           legislation &             criminel &               finance &            \'economie &                market &              conseil  \\ 
                 right &             services &              sentence &        contrevenants &                access &              secteur &             directors &                ouest  \\ 
              services &               accord &                 youth &                peine &                  jobs &               argent &               western &           amendement  \\ 
          negotiations &                 voix &              criminal &                  ans &               economy &              emplois &              election &              comit\'e  \\ 
              chairman &              adopter &                 court &                 juge &              industry &            assurance &               support &          r\'eformiste  \\ 
                public &           r\'eglement &                 issue &              enfants &             financial &          financi\'ere &                 party &               propos  \\ 
                 party &              article &                   law &            important &               billion &              appuyer &                  farm &            important  \\ 
             employees &               retour &             community &                 gens &               support &               cr\'eer &           agriculture &               compte  \\ 
            collective &                 gens &                 right &            tribunaux &                  ovid &                 choc &                clause &                 prix  \\ 
                agreed &              conseil &                reform &                droit &                merger &               acc\'es &                ottawa &                   no  \\ 
                 board &       collectivit\'es &               country &            probl\`eme &           information &            milliards &                    us &         dispositions  \\ 
           arbitration &               postes &               problem &          r\'eformiste &                  size &               propos &                  vote &          information  \\ 
                 grain &                grain &                person &              trait\'e &                 korea &                  pme &                   cwb &               mesure  \\ 
                 order &              tr\'esor &               support &                 faut &             companies &              obtenir &                states &             produits  \\ 
 \end{tabular}
\caption{ The real data (translation) experiment. Topics 5 to 8. } 
\label{tab:2}
\end{table*}
\setlength{\textfloatsep}{9pt}
\begin{table*}[h!] 
\centering 
\begin{tabular}{cc|cc|cc|cc} 
                nisga &                nisga &               pension &              r\'egime &          newfoundland &                terre &                health &               sant\'e \\ 
                treaty &          autochtones &                  plan &             pensions &             amendment &                droit &              research &            recherche  \\ 
            aboriginal &              trait\'e &                  fund &          cotisations &                school &        modifications &                  care &            f\'ed\'eral  \\ 
             agreement &               accord &              benefits &          prestations &             education &            provinces &               federal &            provinces  \\ 
                 right &                droit &                public &             retraite &                 right &               \'ecole &             provinces &                soins  \\ 
                  land &              nations &            investment &               emploi &          constitution &              comit\'e &                budget &               budget  \\ 
               reserve &          britannique &                 money &            assurance &             provinces &           \'education &               billion &              dollars  \\ 
              national &            indiennes &          contribution &       investissement &             committee &         enseignement &                social &             syst\'eme  \\ 
               british &                terre &                   cpp &                fonds &                system &             syst\'eme &                 money &             finances  \\ 
              columbia &             colombie &            retirement &              ann\'ees &                reform &              enfants &                   tax &            transfert  \\ 
                indian &            r\'eserves &                   pay &                  ans &              minority &                 vote &                system &            milliards  \\ 
                 court &                  non &               billion &               argent &        denominational &           amendement &            provincial &              domaine  \\ 
                 party &             affaires &                change &            important &            referendum &         constitution &                  fund &              sociale  \\ 
                   law &        n\'egociations &               liberal &       administration &              children &            religieux &               country &              ann\'ees  \\ 
                native &                bande &           legislation &              dollars &                quebec &         r\'ef\'erendum &                quebec &              maladie  \\ 
                   non &          r\'eformiste &                 board &               propos &               parents &              article &              transfer &            important  \\ 
          constitution &         constitution &            employment &            milliards &              students &          r\'eformiste &                  debt &            programme  \\ 
           development &          application &                   tax &                 gens &                change &              qu\'ebec &               liberal &            lib\'eraux  \\ 
                reform &                 user &                  rate &                 taux &                 party &    constitutionnelle &              services &        environnement  \\ 
           legislation &              gestion &             amendment &                  rpc &              labrador &     confessionnelles &                 issue &            assurance  \\ 
 \end{tabular}
\caption{ The real data (translation) experiment. Topics 9 to 12. } 
\label{tab:3}
\end{table*}
\setlength{\textfloatsep}{9pt}
\begin{table*}[h!] 
\centering 
\begin{tabular}{cc|cc|cc|cc} 
                party &                 pays &                   tax &               agence &                quebec &              qu\'ebec &                 court &              p\^eches  \\ 
               country &            politique &             provinces &            provinces &               federal &          qu\'eb\'ecois &                 right &                droit  \\ 
                 issue &            important &                agency &               revenu &           information &            f\'ed\'eral &             fisheries &                 juge  \\ 
                    us &              comit\'e &               federal &              imp\^ots &             provinces &            provinces &              decision &                cours  \\ 
                debate &            lib\'eraux &               revenue &               fiscal &            protection &           protection &                  fish &                 gens  \\ 
               liberal &          r\'eformiste &             taxpayers &            f\'ed\'eral &                 right &       renseignements &                 issue &            d\'ecision  \\ 
             committee &                 gens &          equalization &        contribuables &           legislation &                droit &                   law &            important  \\ 
                  work &               d\'ebat &                system &                payer &            provincial &            personnel &                  work &                 pays  \\ 
                 order &               accord &              services &                 taxe &                person &               priv\'e &                    us &              trait\'e  \\ 
               support &        d\'emocratique &        accountability &        p\'er\'equation &                   law &            prot\'eger &                 party &         conservateur  \\ 
                reform &          qu\'eb\'ecois &             amendment &               argent &          constitution &        \'electronique &                debate &              r\'egion  \\ 
              election &           r\'eglement &               billion &             services &               privacy &              article &               justice &            probl\`eme  \\ 
                 world &               propos &                 money &             fonction &               country &             commerce &               problem &             supr\'eme  \\ 
                quebec &            coll\'egue &                 party &             modifier &            electronic &          provinciaux &             community &            tribunaux  \\ 
              standing &        parlementaire &            provincial &              article &                 court &                 bloc &               supreme &                 faut  \\ 
              national &              appuyer &                public &           minist\'ere &                  bloc &                  vie &               country &            situation  \\ 
              interest &           opposition &              business &       administration &              students &          application &                  area &             victimes  \\ 
             important &           \'elections &                reform &         d\'eclaration &               section &             citoyens &                  case &              appuyer  \\ 
                 right &                 bloc &                office &                  tps &                 clear &                  non &                 order &               mesure  \\ 
                public &            industrie &               support &          provinciaux &                states &            nationale &            parliament &               trouve  \\ 
 \end{tabular}
\caption{The real data (translation) experiment.  Topics 13 to 16. } 
\label{tab:4}
\end{table*}
\setlength{\textfloatsep}{9pt}
\begin{table*}[t]
\centering  
\begin{tabular}{cc|cc|cc|cc} 
          legislation &            important &              national &            important &                  vote &                 voix &                 water &                  eau  \\ 
                 issue &        environnement &                  area &                 gens &                  yeas &                   no &                 trade &           ressources  \\ 
             amendment &               mesure &                 parks &        environnement &              division &              adopter &             resources &               accord  \\ 
             committee &              enfants &                  work &                parcs &                  nays &                 vote &               country &        environnement  \\ 
               support &              comit\'e &               country &                 pays &                agreed &                  non &             agreement &            important  \\ 
            protection &               propos &                    us &               marine &                deputy &               contre &             provinces &            industrie  \\ 
           information &                 pays &           development &               mesure &                paired &            d\'epenses &              industry &          am\'ericains  \\ 
              industry &              appuyer &               support &               propos &           responsible &               accord &            protection &                 pays  \\ 
             concerned &           protection &             community &            f\'ed\'eral &              treasury &              conseil &                export &            provinces  \\ 
                 right &              article &               federal &               jeunes &               divided &               budget &         environmental &         exportations  \\ 
             important &                droit &                 issue &              appuyer &                 order &              cr\'edit &                    us &             \'echange  \\ 
                change &               accord &           legislation &              ann\'ees &                fiscal &              tr\'esor &            freshwater &         conservateur  \\ 
                 world &                 gens &                  help &            assurance &                amount &                  oui &               federal &      responsabilit\'e  \\ 
                   law &           amendement &               liberal &              gestion &               pleased &                 mise &                 world &                effet  \\ 
              families &              adopter &                 world &         conservateur &                budget &               propos &                 issue &            quantit\'e \\ 
                  work &            industrie &           responsible &               accord &                    ms &                porte &           legislation &              trait\'e  \\ 
              children &                  non &             concerned &              r\'egion &        infrastructure &                  lib &           environment &             commerce  \\ 
                 order &            soci\'et\'e &             committee &            probl\`eme &                 board &             pairs &           responsible &                 unis  \\ 
              national &                porte &               problem &            nationale &               consent &            veuillent &           development &            \'economie  \\ 
                states &                   no &             important &              qu\'ebec &             estimates &                 vice &               culture &                alena  \\ 
 \end{tabular}
\caption{The real data (translation) experiment.  Topics 17 to 20. } 
\label{tab:5}
\end{table*}

\end{document}